\documentclass[11pt,a4paper]{article}
\usepackage{amsmath, amssymb, amsthm}
\usepackage{array}
\usepackage[english]{babel}
\usepackage{booktabs}
\usepackage[round]{natbib}
\usepackage{dsfont}
\usepackage[
	top=25mm,
	bottom=25mm,
	left=25mm,
	right=25mm,
]{geometry}
\usepackage{graphicx}
\usepackage[colorlinks=false,allbordercolors={1 1 1}]{hyperref}
\usepackage{listofitems}
\usepackage{multirow}
\usepackage{subcaption}
\usepackage{url}
\usepackage{enumitem}
\usepackage[table]{xcolor}

\newcommand{\proglang}[1]{\textsf{#1}}
\newcommand{\pkg}[1]{{\fontseries{b}\selectfont #1}}

\graphicspath{{.}}

\setlist[description]{
	leftmargin=2\parindent,
	labelindent=\parindent,
}

\newcommand{\generategridfigures}[1]{
	\setsepchar{,}
	\readlist\functionimages{#1}
	\begin{tabular}{
			@{}m{.07\textwidth}@{}
			>{\centering\arraybackslash}m{.31\textwidth}@{}
			>{\centering\arraybackslash}m{.31\textwidth}@{}
			>{\centering\arraybackslash}m{.31\textwidth}@{}
		}
		& $\boldsymbol{\mathcal{X}}_{\mathrm{F}}$
		& $\boldsymbol{\mathcal{X}}_{\mathrm{sF}}$
		& $\boldsymbol{\mathcal{X}}_{\mathrm{B}}$ \\
		
		$\Upsilon_{\mathrm{lu}}$
		&\includegraphics[width=.31\textwidth]{\functionimages[1]}
		&\includegraphics[width=.31\textwidth]{\functionimages[2]}
		&\includegraphics[width=.31\textwidth]{\functionimages[3]} \\
		
		$\Upsilon_{\mathrm{lb}}$
		&\includegraphics[width=.31\textwidth]{\functionimages[4]}
		&\includegraphics[width=.31\textwidth]{\functionimages[5]}
		&\includegraphics[width=.31\textwidth]{\functionimages[6]} \\
		
		$\Upsilon_{\mathrm{nl}}$
		&\includegraphics[width=.31\textwidth]{\functionimages[7]}
		&\includegraphics[width=.31\textwidth]{\functionimages[8]}
		&\includegraphics[width=.31\textwidth]{\functionimages[9]} \\
		
		$\Upsilon_{\mathrm{d}}$
		&\includegraphics[width=.31\textwidth]{\functionimages[10]}
		&\includegraphics[width=.31\textwidth]{\functionimages[11]}
		&\includegraphics[width=.31\textwidth]{\functionimages[12]} \\
	\end{tabular}
}

\newcommand{\generategridtable}[4]{
	\setsepchar{,}
	\readlist\upsilonlu{#1}
	\readlist\upsilonlb{#2}
	\readlist\upsilonnl{#3}
	\readlist\upsilond{#4}
	\begin{tabular}{clccc}
		\toprule
		& & $\boldsymbol{\mathcal{X}}_{\mathrm{F}}$
		& $\boldsymbol{\mathcal{X}}_{\mathrm{sF}}$
		& $\boldsymbol{\mathcal{X}}_{\mathrm{B}}$ \\
		\cmidrule(lr){3-5}
		\multirow{3}{.05\textwidth}{$\Upsilon_{\mathrm{lu}}$} & FLM & \upsilonlu[1] & \upsilonlu[2] & \upsilonlu[3] \\
		& FKNN & \upsilonlu[4] & \upsilonlu[5] & \upsilonlu[6] \\
		& FNN & \upsilonlu[7] & \upsilonlu[8] & \upsilonlu[9] \\
		[.5em]
		\multirow{3}{.05\textwidth}{$\Upsilon_{\mathrm{lb}}$} & FLM & \upsilonlb[1] & \upsilonlb[2] & \upsilonlb[3] \\
		& FKNN & \upsilonlb[4] & \upsilonlb[5] & \upsilonlb[6] \\
		& FNN & \upsilonlb[7] & \upsilonlb[8] & \upsilonlb[9] \\
		[.5em]
		\multirow{3}{.05\textwidth}{$\Upsilon_{\mathrm{nl}}$} & FLM & \upsilonnl[1] & \upsilonnl[2] & \upsilonnl[3] \\
		& FKNN & \upsilonnl[4] & \upsilonnl[5] & \upsilonnl[6] \\
		& FNN & \upsilonnl[7] & \upsilonnl[8] & \upsilonnl[9] \\
		[.5em]
		\multirow{3}{.05\textwidth}{$\Upsilon_{\mathrm{d}}$} & FLM & \upsilond[1] & \upsilond[2] & \upsilond[3] \\
		& FKNN & \upsilond[4] & \upsilond[5] & \upsilond[6] \\
		& FNN & \upsilond[7] & \upsilond[8] & \upsilond[9] \\
		\bottomrule
	\end{tabular}
}

\DeclareMathOperator{\argmin}{argmin}
\DeclareMathOperator{\argmax}{argmax}

\begin{document}
	
\title{Functional relevance based on \\ the continuous Shapley value}
\author{
	Pedro Delicado and Cristian Pachón-García\\
	{\small Departament d'Estadística i Investigació Operativa} \\
	{\small Universitat Politècnica de Catalunya}
}
\date{\today}
\maketitle

\begin{abstract}
The presence of artificial intelligence (AI) in our society is increasing, which brings with it the need to understand the behavior of AI mechanisms, including machine learning predictive algorithms fed with tabular data, text or images, among others. This work focuses on interpretability of predictive models based on functional data. Designing interpretability methods for functional data models implies working with a set of features whose size is infinite. In the context of scalar on function regression, we propose an interpretability method based on the Shapley value for continuous games, a mathematical formulation that allows for the fair distribution of a global payoff among a continuous set of players. The method is illustrated through a set of experiments with simulated and real data sets. The open source \proglang{Python} package \pkg{ShapleyFDA} is also presented.

\end{abstract}

\section{Introduction}

Technological advances in recent years have affected data analysis significantly. For instance, modern smart devices are able to monitor health indicators sampling data several times per minute. This way of obtaining data leads to treat them as if they were functions depending on a continuous argument (time, wavelength, etc.). {\em Functional data analysis} \citep{Ramsay2005}, or FDA for short, is a branch of statistics whose theoretical foundations come from functional analysis (see also \citealp{Ferraty2006,Horvath2012,Kokoszka2017,Ciprian2024}). In particular, prediction (understood in a broad sense) is a relevant problem in FDA. To fix ideas, given a functional explanatory variable (or functional feature) $\boldsymbol{\mathcal{X}}(t)$, $t\in I=[a,b]\subset \mathbb{R}$, and a scalar response variable (or target) $Y$, one aims to find the best mapping between the functional feature and the target. This problem is known as {\em scalar-on-function regression}, and it has been addressed not only with statistical prediction methods (see, for example, \citealp{Reiss:2017,Gertheiss2024}), but also with machine learning algorithms (see, for instance, \citealp{Yao2021,Aniruddha2023,Gertheiss2024}). We focus on regression ($Y$ continuous) but classification ($Y$ binary or categorical) is handled in a similar way.

Machine learning models, even in the standard case of multivariate features, have enormous flexibility to encode the relationship between the explanatory variables and the response. This typically leads to difficulties in understanding exactly how machine learning algorithms work, justifying the use of the term ``black box" to describe them. In the last two decades, a powerful research line (known as interpretable machine learning) has been developed to provide interpretability tools to algorithmic models. There is a considerable number of review papers (see, for instance, \citealp{Barredo2020}) and three monographs on this topic: \citet{Biecek2021}, \citet{Masis2021}, and \citet{Molnar2022}.

In this work we present a methodology, based on game theory, which provides interpretability to scalar-on-function prediction models. Our proposal can explain any kind of model (it is {\em agnostic} to the underlying prediction model) and it provides a {\em global} explanation (it shows how important each value of the functional explanatory variable is in the prediction process, taking into account all the observed data). To the best of our knowledge, this is the first work that addresses global agnostic interpretability for prediction models in the context of FDA. In fact, it is noteworthy that an examination of the intersection between functional data analysis and interpretable machine learning reveals a limited corpus of existing literature: only two works can be found, \citet{James2009}, which is not agnostic to the model, and \citet{Carrizosa2024}, which is not global.

The structure of this paper is as follows. Section \ref{sec:game_theory} shows the connection between the Shapley value for finite games and interpretability for multiple regression, and introduces the Shapley value for continuous games. Based on it, our interpretability framework is detailed in Section \ref{sec:shapley_fda}, while Section \ref{sec:package} introduces the accompanying open source \proglang{Python} package \citep{PythonCoreTeam}. In order to study and analyze this proposal, we conduct a set of simulations in Section \ref{sec:experiments}, which additionally includes a real data example. Conclusions are provided in Section \ref{sec:conclusions}. Some appendixes are available as supplementary material: Appendix \ref{sec:iml} offers a brief introduction to interpretable machine learning, and Appendix \ref{sec:results_simulation_study} includes extra outputs from simulations.

\section{Interpretability based on game theory}
\label{sec:game_theory}

This section establishes the connection between the Shapley value for finite games, a concept from game theory, and interpretability for multiple regression. We also introduce the Shapley when the game is continuous, the framework for our proposal in Section \ref{sec:shapley_fda}.

\subsection{Finite games}
\label{sec:finite_games}

In this Section we follow \citet{Winter2002} to present the Shapley value. The seminal work was published by \citet{Shapley1953}. Given a set of players $N=\{1, \dots, n\}$, a {\em game} is a function $\nu: 2^N \rightarrow \mathbb{R}^+$, where $2^N$ is the set of all subsets of $N$ and $\nu(\emptyset) = 0$. The mapping $\nu$ is known as the {\em payoff function} and $\nu(S)$ is interpreted as the payoff the coalition $S$ receives for having played that game in a cooperative way. When all the players cooperate, the total payoff is $\nu(N)$. The relevant question in cooperative game theory is to find a fair distribution of $\nu(N)$ among the $n$ players, or rephrased differently, to determine the relevance (the value) of each player in the overall coalition.

Let $\mathcal{Q}_N$ be the set of all games $\nu$ with $n$ players. A {\em value} is an operator $\varphi: \mathcal{Q}_N \rightarrow \left(\mathbb{R}^{+}\right)^n$ that assigns to each game a vector of length $n$, $\varphi(\nu) = \boldsymbol{\varphi}_{\nu} \in \left(\mathbb{R}^{+}\right)^n$, where the $i$-th component of the vector, $\varphi_{\nu, i}$ (or $\varphi_i$ whenever the game can be omitted), represents the value of the player $i$ when playing game $\nu$. \citet{Shapley1953} defines a set of axioms (or desirable properties) for values:
\begin{description}
	\item[Efficiency.] The sum of all values must equal the total payoff: $\sum_{i=1}^n \varphi_i = \nu(N)$.
	
	\item[Symmetry.] Players $i, i' \in N$ are {\em symmetric} with respect to the game $\nu$ if for each $S \subset N$ such that $i, i' \notin S$, $\nu(S \cup \{i\}) = \nu(S \cup \{i'\})$. 
	If this is the case then $\varphi_i = \varphi_{i'}$.
	
	\item[Dummy players.] Player $i$ is a {\em dummy player} (with respect to the game $\nu$) if for every $S \subset N$, $\nu(S \cup \{i\}) - \nu(S) = 0$. If player $i$ is a dummy player, then $\varphi_i = 0$.
	
	\item[Additivity.] $\varphi(\nu + \omega) = \varphi(\nu) + \varphi(\omega)$, where $\nu$ and $\omega$ are games, and game $\nu + \omega$ is defined as $(\nu + \omega)(S) = \nu(S) + \omega(S)$.
\end{description}
\citet{Shapley1953} proves that there is a unique value (now known as the {\em Shapley value} of the game) which satisfies the previous axioms, and that it is given by
\begin{equation}
    \label{eq:shapley_finite}
    \varphi_{\nu, i} = \frac{1}{n!} \sum_{\pi \in \Pi}\big[ \nu(p_{\pi}^i \cup \{i\}) - \nu(p_{\pi}^i)\big],
\end{equation}
where $\Pi$ is the set of all permutations of the set $N$, $\pi$ is a permutation and $p_{\pi}^i\!=\!\{j\!\!:\! \pi(i)\! > \!\pi(j)\}\!$ is the set of players preceding player $i$ in permutation $\pi$. The quantity $\big[\nu(p_{\pi}^i \cup \{i\}) - \nu(p_{\pi}^i)\big]$ is the marginal contribution of player $i$ to the coalition $p_{\pi}^i$ and its Shapley value, $\varphi_{\nu, i}$, is the average of these marginal contributions over the possible permutations of $N$.

\subsection{Interpretability based on finite games}
\label{sec:iml_finite_games}

Consider a multiple linear regression model with $m$ individuals and $p$ explanatory variables. Let $\mathbf{x}_1, \ldots, \mathbf{x}_p$ and $\mathbf{y}$ be, respectively, the vectors of predictors' values and responses, all in $\mathbb{R}^m$. Let $\bar{y} = 1/m \sum_{j=1}^m y_j$ and let $\hat{y}_j$ be the $j$-th fitted value by ordinary least squares (OLS). The coefficient of determination, $R^2= 1 - \sum_{j=1}^m (y_j - \hat{y}_j)^2/\sum_{j=1}^m (y_j - \bar{y}_j)^2$, is commonly used to measure the overall quality of the estimated model. When the $p$ explanatory variables are uncorrelated, $R^2 = \sum_{i=1}^p R_i^2$, where $R^2_i$ is the coefficient of determination in the simple linear regression of $\mathbf{y}$ against the $i$-th explanatory variable $\mathbf{x}_i$ fitted by OLS. Therefore, $R^2_i$ is the contribution of $\mathbf{x}_i$ to the global quality measure $R^2$, and it is a good measure of the relevance of $\mathbf{x}_i$ in the model. See, for instance, \citet{Gromping2009}.

The preceding decomposition of $R^2$ is not applicable when the explanatory variables are correlated. \citet{Lipovetsky2001} propose an alternative decomposition based on the Shapley value. Specifically, the authors propose to consider the $p$ explanatory variables as the set of players and the game $\nu$ as the coefficient of determination $R^2$. Consequently, $\nu(S)$ is the coefficient of determination $R^2(S)$ in the regression of $\mathbf{y}$ against the variables belonging to $S$ fitted by OLS. Therefore, the Shapley value of this game is a fair distribution of the total $R^2$ among the $p$ predictors: $R^2 = \sum_{i=1}^p \varphi_{\nu, i}$, and $\varphi_{\nu, i}$ measures the importance of the $i$-th regressor in the model. Given that the exact computation of Shapley values is quite time intensive, the authors suggest to average over a moderate number of random permutations of the explanatory variables. \citet{Feldman2005} extends the use of the Shapley value to parametric statistical or econometric models, as a measure of the relative importance of each variable. \citet{Cohen2007} propose to use the Shapley value as global measure of variable relevance in classification problems using any prediction model (or algorithm). They propose to use the accuracy in a test set as the game $\nu$.

\subsection{Continuous games}
\label{sec:continuous_games}

The primary objective of this work is to incorporate interpretability within the context of FDA. Therefore, it is considered to extend the previous relevance measure based on the Shapley value to prediction models with scalar response and functional regressor. In this context the set of players $\mathcal{X}(t)$, $t\in I$, is infinite, or, equivalently, the game becomes continuous. Thus, the Shapley value must be defined for {\em continuous games} (or {\em games with a continuum of players}: \citealp{Shapley1961,Aumann1964}). There exists a collection of publications devoted to this topic (\citealp{Kannai1966,Aumann1974,Hart1988,Neyman1994}; among others). Obtaining the Shapley value for a continuous game requires a more sophisticated mathematical framework than for a finite game.

Let $I = [0,1]$ and $\mathcal{B}=\mathcal{B}(I)$ be the associated Borel $\sigma$-algebra. A (continuous) {\em game} is a function $\nu: \mathcal{B} \rightarrow \mathbb{R}^+$ such that $\nu(\emptyset) = 0$. $I$ is called {\em the set of players}, $\mathcal{B}$ {\em the set of coalitions}, $(I, \mathcal{B})$ {\em the space of players} and $\nu$ the {\em payoff function}. Although we work with $I = [0,1]$, the forthcoming development is valid for any interval $[a,b] \subset \mathbb{R}$.

A game is said {\em monotonic} if $T \subset S$ implies $\nu(T) \leq \nu(S)$. Let $Q$ be a collection of monotonic games in $(I, \mathcal{B})$. $Q$ is a {\em linear space} over $(I, \mathcal{B})$ if $\nu_1 + \nu_2 \in Q$ for any $\nu_1$ and $\nu_2$, both in $Q$. An {\em automorphism} in $(I,\mathcal{B})$ is a one-to-one measurable function $\theta$ from $I$ onto $I$ with $\theta^{-1}$ measurable. Let $\mathcal{G}$ be the group of automorphisms of $(I, \mathcal{B})$. Given a game $\nu$, an automorphism $\theta\in\mathcal{G}$ defines a new game from $\nu$: $(\theta_*\nu)(S)=\nu(\theta(S))$ for all $S\in \mathcal{B}$. $Q$ is {\em symmetric} if $\theta_*\nu\in Q$ for all $\nu\in Q$ and all $\theta\in\mathcal{G}$. A measure $\mu$ on $(I, \mathcal{B})$ is a monotonic game with the property of $\sigma$-additivity: for all countable collections $\{E_j\}_{j \geq 1}$ of pairwise disjoint sets in $\mathcal{B}$, $\mu\left(\cup_{j \geq 1}E_j\right) = \sum_{j \geq 1}\mu(E_j)$. Let $\mathcal{M}$ be the set of measures defined on $(I,\mathcal{B})$. Note that $\mathcal{M}$ itself is a symmetric and linear subset of monotonic games.

A {\em value} on a linear symmetric space of monotonic games $Q$ is a mapping $\psi: Q \rightarrow \mathcal{M}$ satisfying the following properties \citep[page 16]{Aumann1974}:
\begin{description}
    \setlength{\itemindent}{0pt}
	\item[Efficiency.] For all $\nu\in Q$, $[\psi(\nu)](I) = \nu(I)$.
	
	\item[Symmetry.] For all $\nu\in Q$ and for all $\theta \in \mathcal{G}$, $\theta_*\psi(\nu) = \psi(\theta_*\nu)$.
	
	\item[Linearity.] For all $\nu_1, \nu_2 \in Q$ and for all $\alpha, \beta \in \mathbb{R}^+$, $\psi(\alpha \nu_1 + \beta \nu_2) = \alpha\psi(\nu_1) + \beta \psi (\nu_2)$.
\end{description}

While the concepts of efficiency and linearity are relatively straightforward to interpret, the notion of symmetry is more abstract. According to \citet{Aumann1974}, it means that {\em ``the value does not depend on how the players are named''}. In essence, this property states that the value of $\nu$ is preserved by automorphisms.

In \citet{Aumann1974}, the authors prove that, under certain assumptions on the linear symmetric space $Q$ of monotonic games defined on $(I, \mathcal{B})$, there exists a unique value $\psi$ satisfying the efficiency, symmetry and linearity properties. Nevertheless, deriving a formula in the continuous games context is not as straightforward as it is in the finite case. The literature referenced at the beginning of this section presents two distinct methodologies for addressing this issue. The first one, known as {\em the axiomatic approach}, considers only a very specific subset of games for which a closed formula can be found. The so called {\em vector measure games} are an example of this situation. Let $\tau=(\tau_1,\ldots,\tau_K)$ be a vector of $K$ non-atomic measures defined on $\mathcal{B}$, with $\tau_k(I)\ne 0$. Let $\rho$ be a real valued continuously differentiable function defined on $\mathbb{R}^K$ with $\rho(0,\ldots,0)=0$. A vector measure game has the form $\nu=\rho \circ \tau$. That is, $\nu(S)=\rho(\tau_1(S),\ldots,\tau_K(S))$, so the payoff of coalition $S$ only depends on a finite number $K$ of measures of $S$. It can be proved that there exist an efficient, symmetric, and linear value function $\psi$ for these games given by
\begin{equation*}
    \psi(\rho \circ \tau)(S) = \sum_{k=1}^K \tau_i(S) \int_0^1 \frac{\partial \rho}{\partial x_k}(t\tau_1(I),\ldots,t\tau_K(I))dt
\end{equation*}
which is called {\em the diagonal formula}. Further details are provided in Section 3.2 of \citet{Neyman1994}. This approach, however, is not generally applicable in the context of the present paper, as it is shown in Section \ref{sec:No_vector_measure_game}.

The second approach, {\em the asymptotic approach}, considers a given continuous game as the limit of a sequence of finite games. This, in turns, allows computing the sequence of (finite) Shapley values. Therefore, the limit value for that sequence is the Shapley value for the continuous game. We consider to use the asymptotic approach because it is not as restrictive as the axiomatic approach. Our development is based on \citet{Kannai1966} and \citet{Neyman1994}. Let $\nu: \mathcal{B} \to \mathbb{R}^+$ be a game, let $\mathcal{I} = \{I_1, \dots, I_q\}\subset \mathcal{B}$ be a partition of $I$, let $\mathcal{P}$ be the $\sigma$-algebra generated by $\mathcal{I}$, and let $\nu_{\mathcal{P}}$ be the game restricted to $\mathcal{P}$, that is, $\nu_{\mathcal{P}}: \mathcal{P} \rightarrow \mathbb{R}^+$ and $\nu_{\mathcal{P}}(S) = \nu(S)$ for all $S \in \mathcal{P}$. As $\nu_{\mathcal{P}}$ is a finite game, the Shapley value described in Section \ref{sec:finite_games}, $\varphi(\nu_{\mathcal{P}})$, can be obtained.

In order to define the limit value, we need to consider a sequence of finer and finer partitions. Let $S \in \mathcal{B}$, a $S$-{\em admissible} sequence is an increasing sequence of subalgebras $(\mathcal{P}_j)_{j \in \mathbb{N}}$ with $S \in \mathcal{P}_1 \subset \mathcal{P}_2 \subset\cdots \subset \mathcal{P}_k \subset \cdots$ such that $\cup_{j \in \mathbb{N}} \mathcal{P}_j$ generates $\mathcal{B}$. $\psi(\nu)$ is {\em the asymptotic value of } $\nu$ if and only if for every $S \in \mathcal{B}$ and any $S$-admissible sequence $(\mathcal{P}_j)_{j \in \mathbb{N}}$, $\lim_{j\to\infty} [\varphi(\nu_{\mathcal{P}_j})](S) = [\psi(\nu)](S)$. Under certain conditions, it can be proved that the limit exists, it is unique and it satisfies the properties previously described \citep{Neyman1994}. Therefore, under those conditions, the value for a continuous game can be obtained as the limit of the Shapley value for finite games.

\section{Relevance based on the continuous Shapley value}
\label{sec:shapley_fda}

In this section, we present a framework for defining a relevance function for prediction models with scalar response $Y$ and a functional predictor $\boldsymbol{\mathcal{X}}$ with trajectories in $L^2(I)=\{g:I\rightarrow \mathbb{R}: \int_I g(t)^2dt<\infty\}$, where $I=[0,1]$ (for simplicity, we use $[0,1]$ instead a generic interval $[a,b]$). The goal is to rank the points $t$ in $I=[0,1]$ according to their importance when predicting the response variable using a particular prediction model. We make use of the asymptotic approach described in Section \ref{sec:continuous_games}.

Let $\boldsymbol{\mathcal{X}}$ and $Y$ have a certain joint probability distribution, from which a test data set is available: $(\mathcal{X}_j,y_j)$, $j \in\{1,\ldots, m\}$, independently drawn from $(\boldsymbol{\mathcal{X}},Y)$. Additionally, it is assumed that one has access to an already trained prediction model, $f: L^2(I) \rightarrow \mathbb{R}$, and that the data used to train such a model are independent of the test data.

One of the key elements is how to define the game $\nu: \mathcal{B} \to \mathbb{R}^+$ which will allow us to find the {\em Shapley value relevance function}. For $S \in \mathcal{B}$, roughly speaking $\nu(S)$ should be the proportion of variability of $\mathbf{y} = (y_1, \ldots, y_m)^\intercal$ in the test sample explained by the prediction model $f$ when only the information of $S$ is considered.

A first attempt to compute $\nu(S)$ could be to retrain the model only considering the points $t\in S$, but this strategy would eventually be computationally very expensive as it would need to be performed for each $S \in \mathcal{B}$. This approach has certain similarities to the {\em leave-one-covariate-out} (LOCO) method for feature relevance in prediction models with a finite number of predictors (for details, see Appendix \ref{sec:iml} in the supplementary material). \citet{Delicado2023} propose a computationally cheaper alternative to LOCO: instead of leaving one covariate out, they replace it by its {\em ghost variable}, which is its conditional expectation given the rest of covariates. 
A related approach is employed in this work.

Specifically, we propose the creation of a new functional data set in which the data from $t\in S$ are retained while the remaining data are inferred from $S$. The new data set is created using the conditional expectation, that is,
\begin{equation}
	\label{eq:x_recumputed}
	\tilde{\mathcal{X}}^S_j(t) = \mathcal{X}_j(t) \cdot \mathds{1}_S(t) + \dot{\mathcal{X}}_j(t) \cdot \mathds{1}_{S^\mathsf{c}}(t),
\end{equation}
with
\begin{equation*}
    \dot{\mathcal{X}}_j(t) = \mathbb{E}\big( \boldsymbol{\mathcal{X}}(t)\mid \{\boldsymbol{\mathcal{X}}(u)=\mathcal{X}_j(u): u \in S\}\big),
\end{equation*}
where $S^\mathsf{c} = I \backslash S$ and $j \in \{1, \ldots, m\}$. This way, the functions $\tilde{\mathcal{X}}^S_j(t)$ are evaluated in all $t \in I$ and they can be used as arguments of the already trained prediction model $f$. Conditional expectations are estimated under the assumption that $\boldsymbol{\mathcal{X}}$ is a Gaussian process. Details are given in Section \ref{sec:cond_expectation}, along with a discussion of its parallels with the approach of \citet{Kneip2020} to optimal reconstruction of partially observed functional data.

Next, we define $\tilde{y}_j^S = f(\tilde{\mathcal{X}}^S_j)$, $j \in \{1,\dots, m\}$, and $\nu(S)$ is defined as the coefficient of determination, computed as
\begin{equation}
    \label{eq:payoff_R2}
	\nu(S)=\tilde{R}^2(S)=1-\frac{\sum_{j=1}^m (y_j-\tilde{y}_j^S)^2}{\sum_{j=1}^m (y_j-\bar{y})^2},
\end{equation}
where $\bar{y}$ is the average of the target (or response) in the test set. Observe that $\tilde{R}^2(I)$ coincides with the standard $R^2(I)$ because $\tilde{\mathcal{X}}^I_j=\mathcal{X}_j$ in Equation (\ref{eq:x_recumputed}). 

\begin{figure}
	\centering
	\begin{subfigure}[t]{.32\textwidth}
		\centering
		\includegraphics[width=\textwidth]{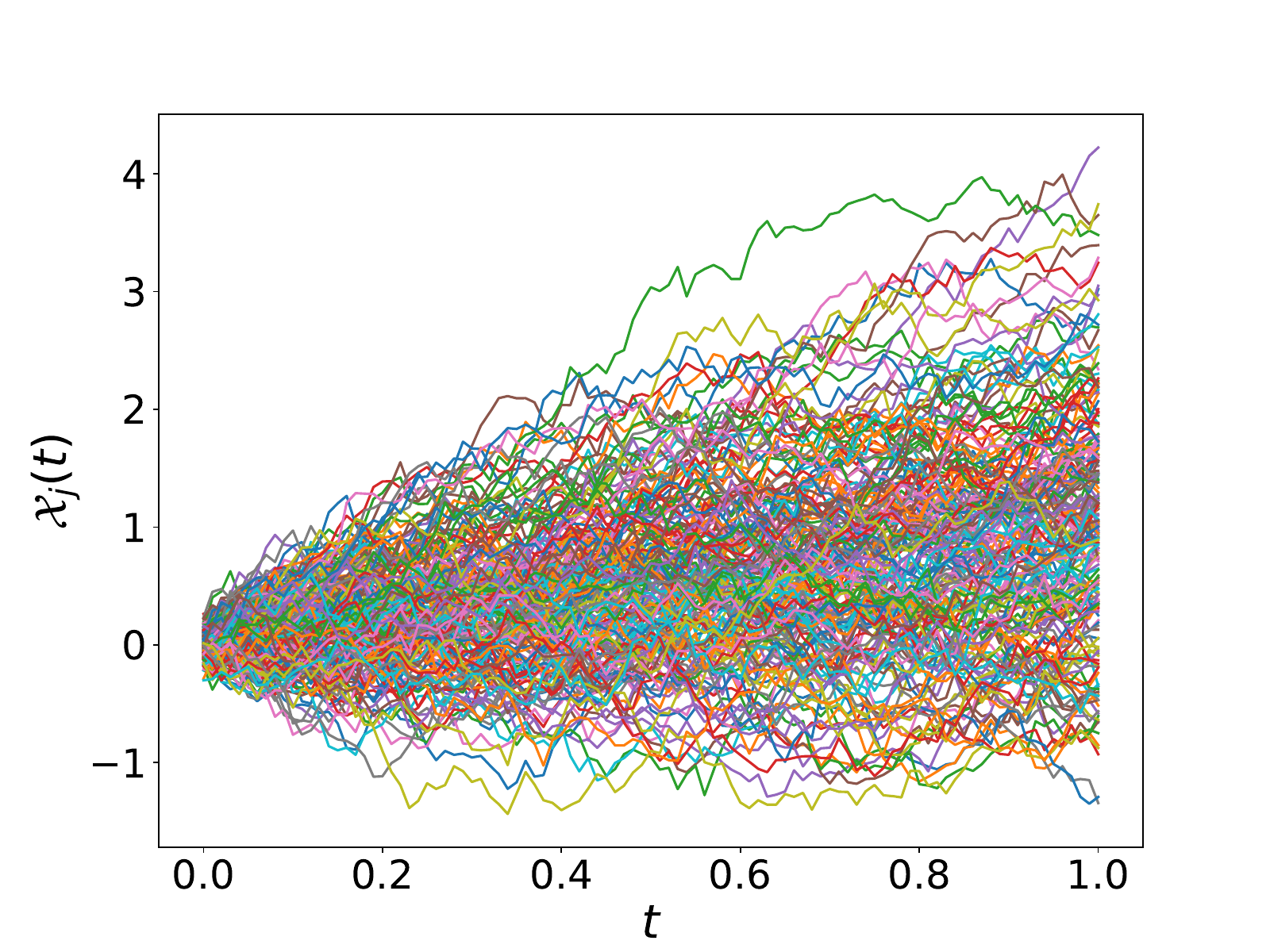}
	\end{subfigure}
	\hfill
	\begin{subfigure}[t]{.32\textwidth}
		\centering
		\includegraphics[width=\textwidth]{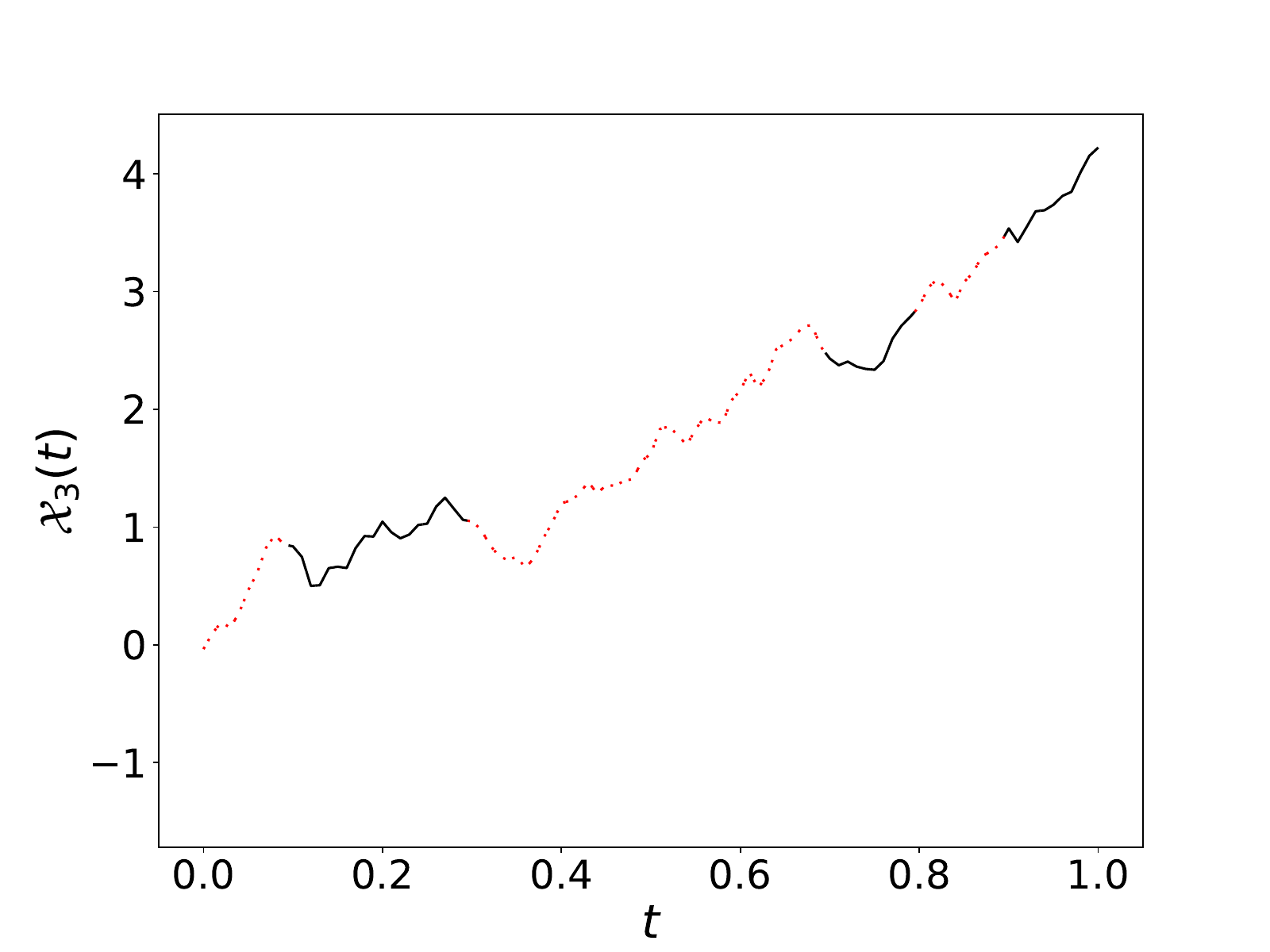}
	\end{subfigure}
	\hfill
	\begin{subfigure}[t]{.32\textwidth}
		\centering
		\includegraphics[width=\textwidth]{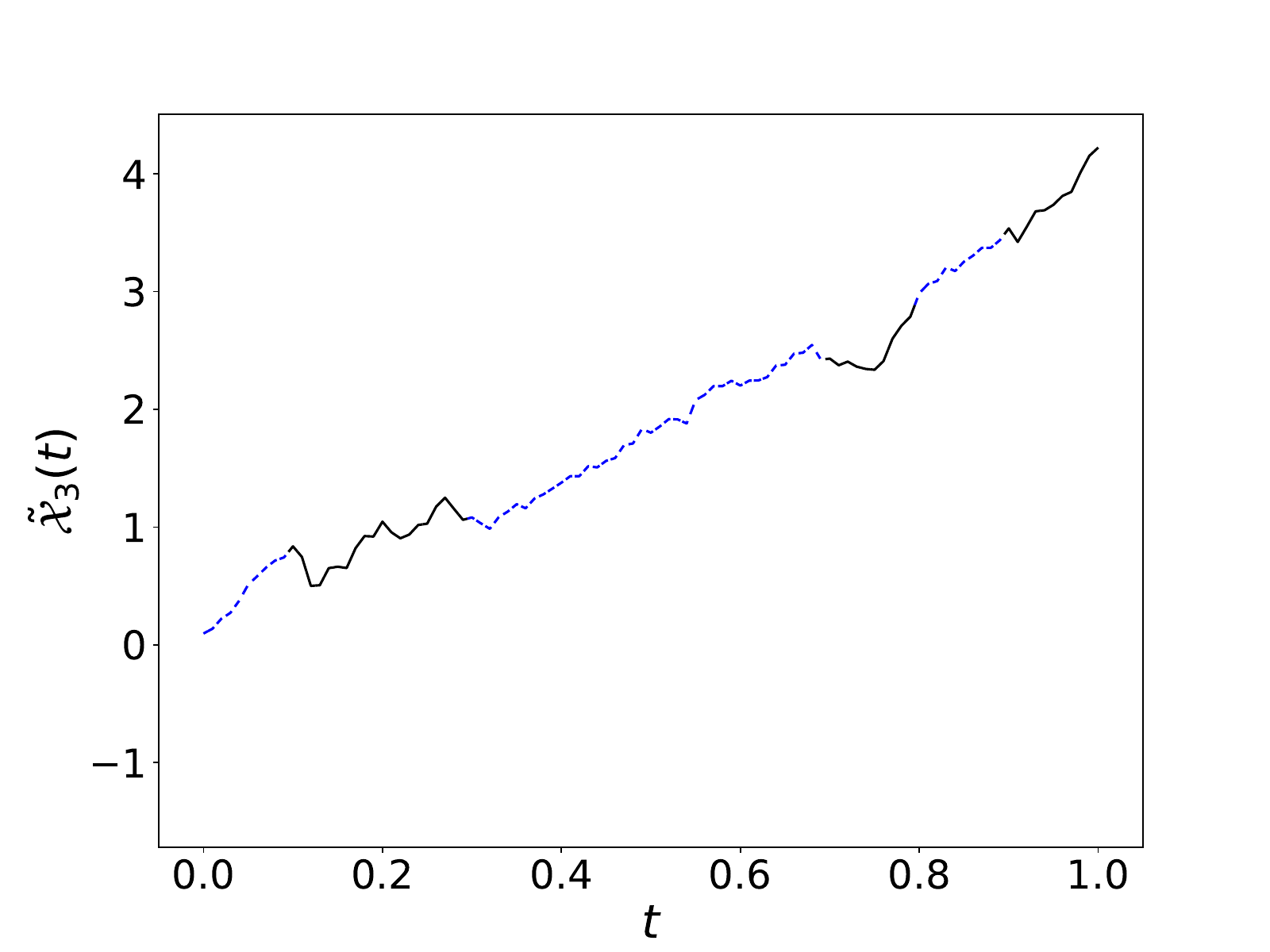}
	\end{subfigure}
	
	\caption{Example of a reconstructed function. (Left) Functional data set. (Middle) One functional data. Solid lines indicate intervals in $S$ where the functional data is known, while dotted lines correspond to $S^{\mathsf{c}}$ where the functional data is unknown. (Right) Reconstructed function. The values of the functional data for the intervals in $S^{\mathsf{c}}$ (dashed lines) are computed as their conditional expectation given the values at the intervals in $S$ (solid lines).}
	\label{fig:cond_expectation}
\end{figure}

Figure \ref{fig:cond_expectation} shows an example of how the new data set is obtained. In the left panel, there is a data set with 100 functions $\mathcal{X}_j(t)$ $j \in \{1, \ldots, m=100\}$. Details on how this data set was generated are given in Section \ref{sec:exp_simulated_data}. The middle plot is one of these functions. The intervals that are part of $S$ are shown in solid lines, while those that are part of $S^\mathsf{c}$ are shown in dotted lines. The right side of the figure shows the reconstructed function $\tilde{\mathcal{X}}^S_j(t)$, where the dashed lines indicates those intervals for which conditional expectation was used.

Recall that our goal is to make use of the asymptotic approach to compute the Shapley value of the game $\nu$. Therefore, let $\mathcal{I} = \{I_1, \ldots, I_n\}$ be a partition of $I = [0,1]$, where $I_i = [a_i, b_i)$, for $i \in \{1, \ldots, n-1\}$, $I_n =[a_n, b_n]$, $a_1=0$, $b_n=1$ and $b_{i-1}=a_i<b_i$ for $i \in \{2, \ldots, n\}$. Let $\mathcal{P}$ be the $\sigma$-algebra generated by $\mathcal{I}$. In the context of game theory, $\mathcal{I}$ (or, equivalently, $N = \{1, \ldots, n\}$) can be thought of as the set of players. Therefore, we define the finite game $\nu_{\mathcal{P}} : \mathcal{P} \to \mathbb{R}^+$ as the previous game $\nu$ restricted to $\mathcal{P}$.

The objective is to compute the Shapley value for a given subset $I_i$ (player $i$) using the formula stated in Equation (\ref{eq:shapley_finite}). Let $\Pi$ the set of all permutations of $\mathcal{I}$, let $\pi \in \Pi$ and let $p^i_{\pi}$ the set of players preceding $i$ in $\pi$. By using Equation (\ref{eq:x_recumputed}) it is not necessary to retrain a model for each different permutation.

In this context, the role of $S$ is assumed by $p_\pi^i$ (and then $S^\mathsf{c}$ is $\mathcal{I} \backslash p_\pi^i$). This, in turn, allows computing $\nu_{\mathcal{P}}(p_\pi^i)$. The same line of reasoning can be applied to $p_\pi^i \cup I_i$. Using the Shapley value for finite games, the relevance of the interval $I_i$ is given by
\begin{equation}
	\label{eq:shapley_cont_formula}
	\varphi_{\nu, i} = \frac{1}{n!} \sum_{\pi \in \Pi}\big[ \nu_{\mathcal{P}}(p_{\pi}^i \cup I_i) - \nu_{\mathcal{P}}(p_{\pi}^i)\big] = \frac{1}{n!} \sum_{\pi \in \Pi}\big[ \tilde{R}^2(p_{\pi}^i \cup I_i) - \tilde{R}^2(p_{\pi}^i)\big].
\end{equation}
In practice, as the size of the set of all permutations $\Pi$ could be extremely large, a random subset of permutation $\Pi_0 \subset \Pi$ is employed.

Applying the previous formula to each interval leads to obtain an $n$ dimensional vector $\boldsymbol{\varphi} = (\varphi_{\nu, 1}, \ldots, \varphi_{\nu, n})^\intercal$. So the {\em Shapley value relevance function} is the density function of the histogram-type measure defined by vector $\boldsymbol{\varphi}$:
\begin{equation*}
    \mathcal{R}_f(t)= \sum_{i=1}^n \frac{\varphi_{\nu, i}}{b_i - a_i} \mathds{1}_{I_i}(t).
\end{equation*}
The Shapley value relevance function can also be reported as the polygonal function defined by the midpoints of the steps in $\mathcal{R}_f(t)$ and its boundary values.

\begin{figure}
	\centering
	\begin{subfigure}[t]{.48\textwidth}
		\centering
		\includegraphics[width=\textwidth]{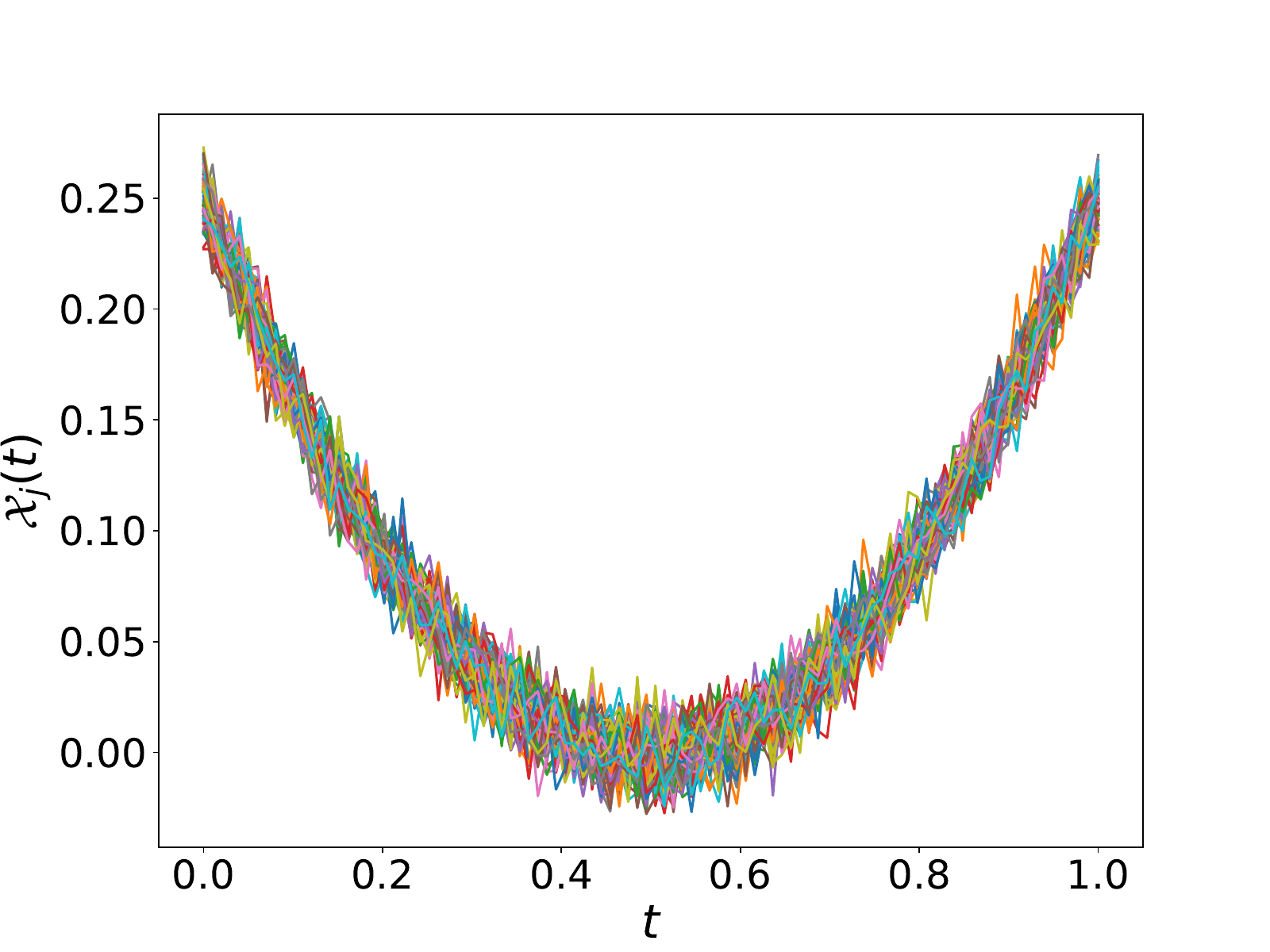}
		\caption{Functional data set that is generated according to $\mathcal{X}(t) = (t - 0.5)^2 + \varepsilon(t)$.}
		\label{fig:toy_functional_covariates}
	\end{subfigure}
	\hfill
	\begin{subfigure}[t]{.48\textwidth}
		\centering
		\includegraphics[width=\textwidth]{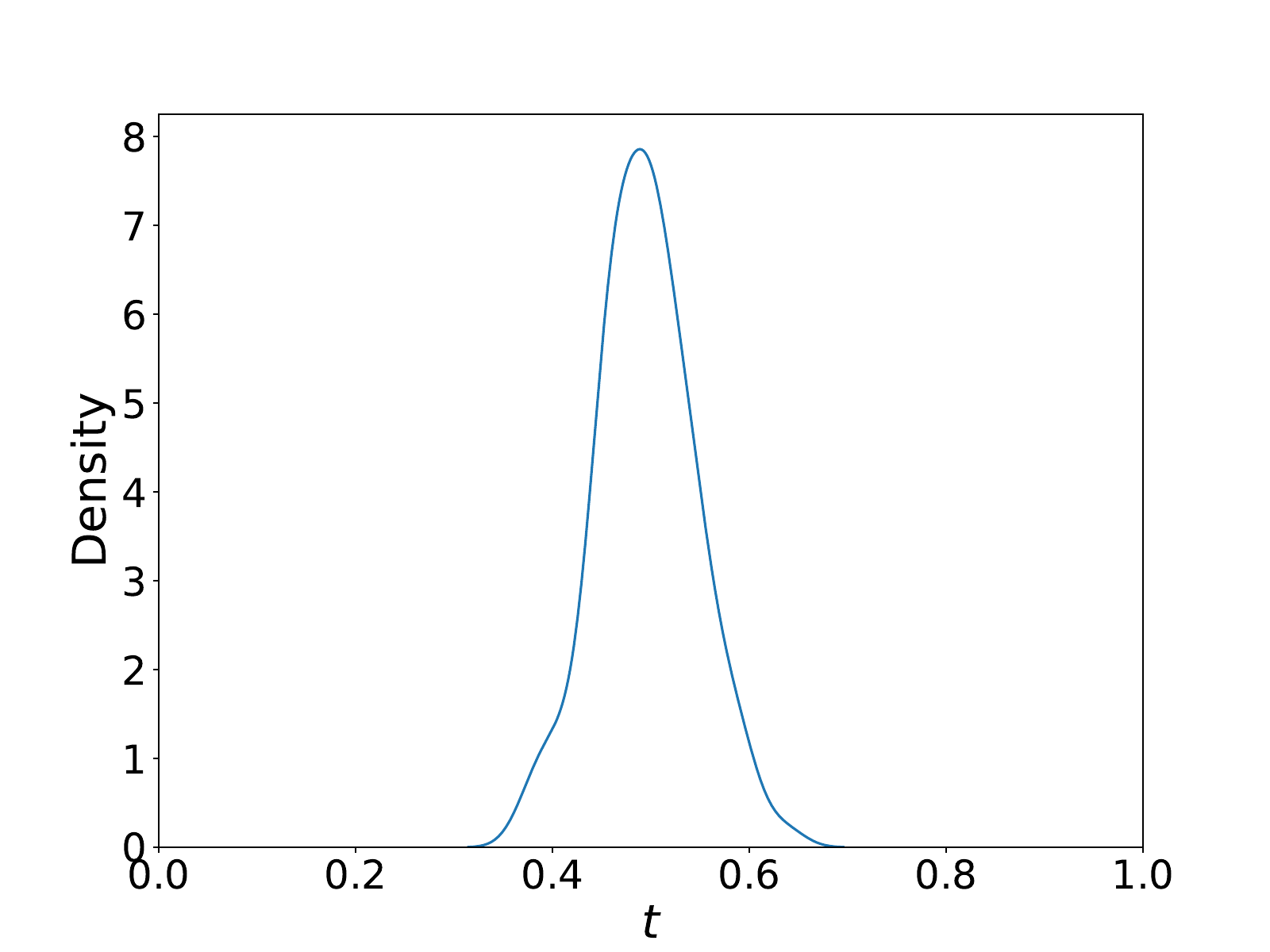}
		\caption{Estimated density of $\argmin[\mathcal{X}(t)]$. 
			The target variable is generated as $Y = \min_{t \in I}[\mathcal{X}(t)]$.}
		\label{fig:toy_density}
	\end{subfigure}
	\hfill
	\begin{subfigure}[t]{.48\textwidth}
		\centering
		\includegraphics[width=\textwidth]{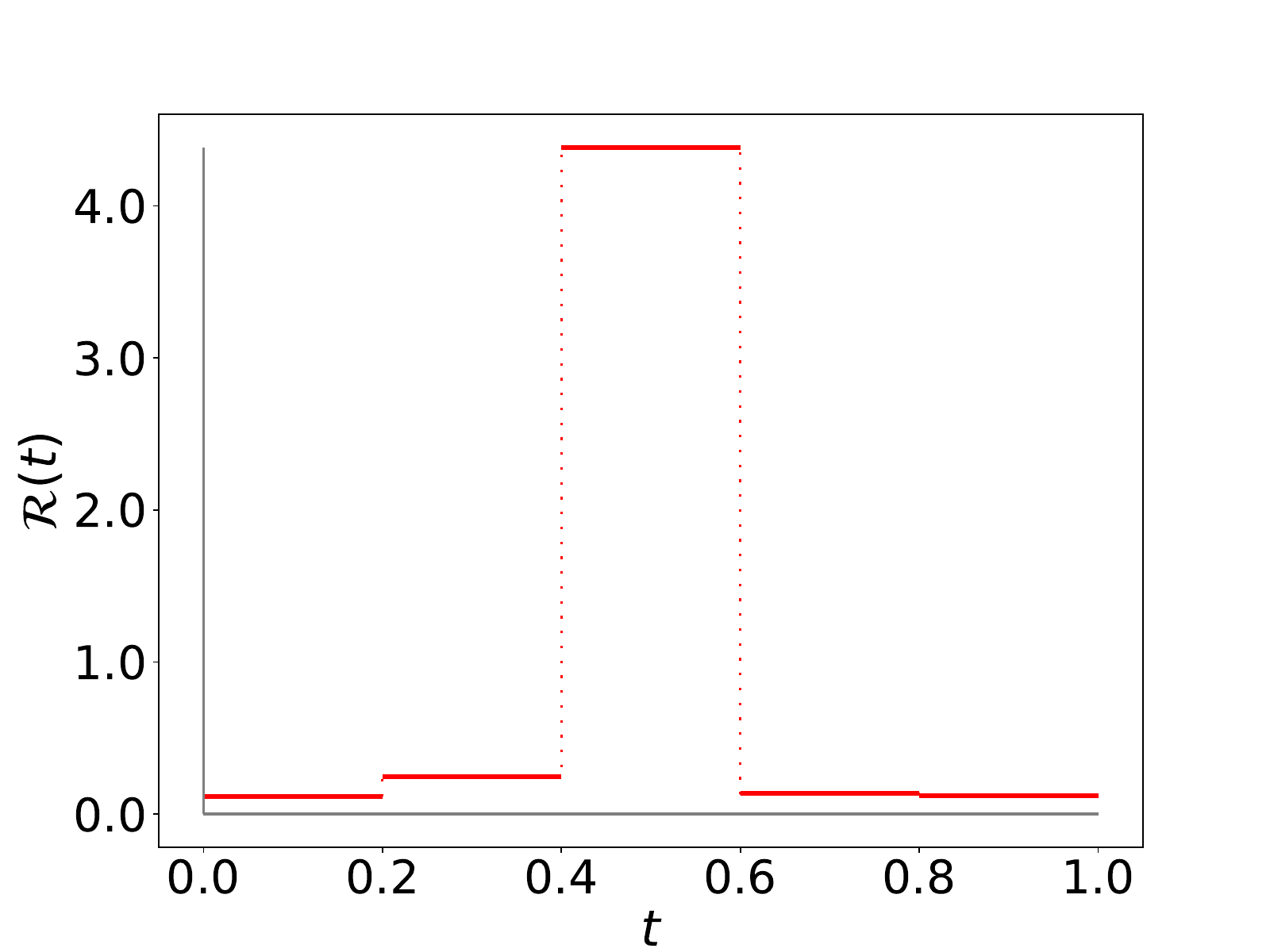}
		\caption{Histogram-type Shapley value relevance function.}
		\label{fig:toy_histogram}
	\end{subfigure}
	\hfill
	\begin{subfigure}[t]{.48\textwidth}
		\centering
		\includegraphics[width=\textwidth]{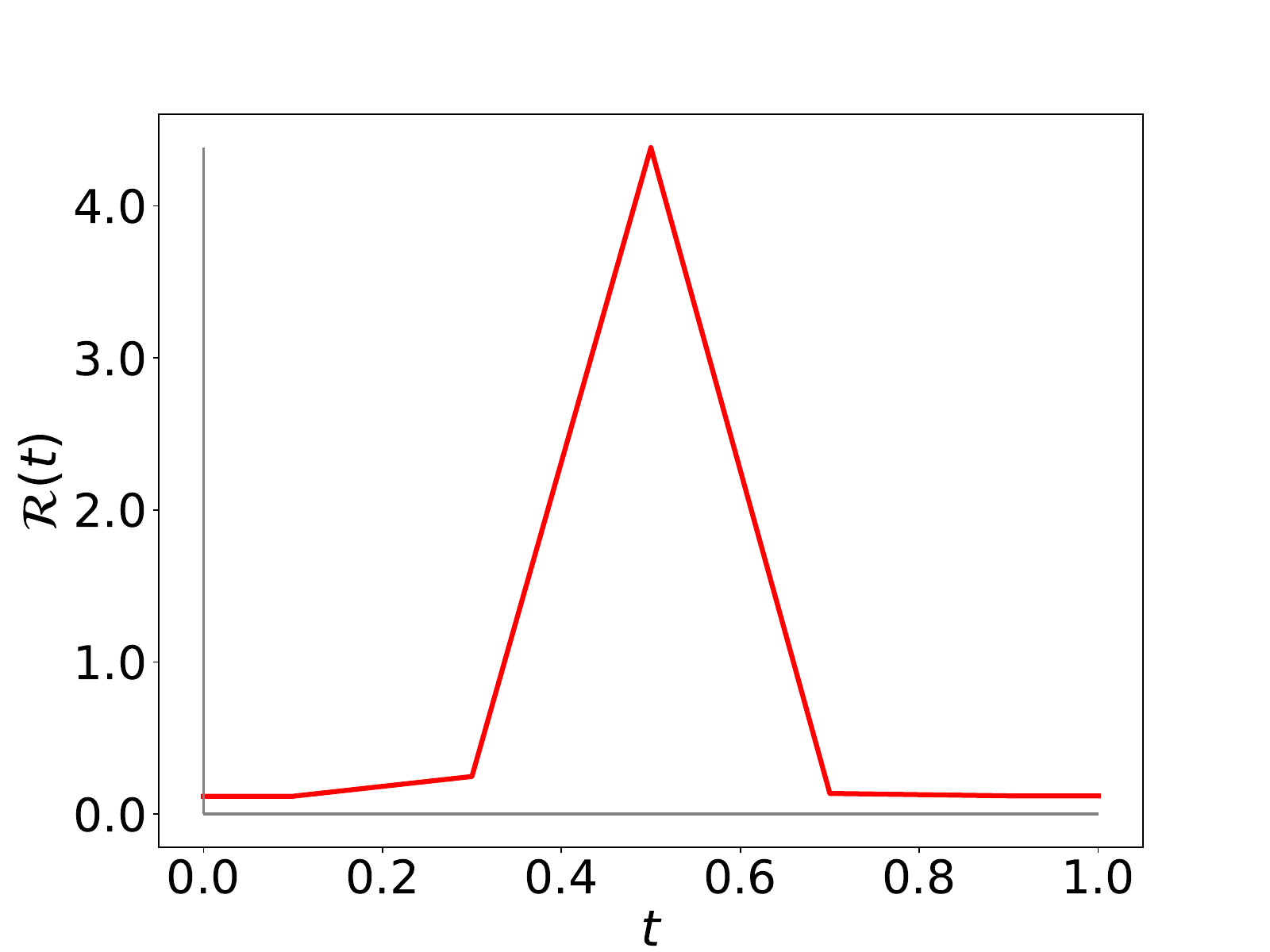}
		\caption{Polygonal-type Shapley value relevance function.}
		\label{fig:toy_polygonal}
	\end{subfigure}
	
	\caption{Illustrative example of the Shapley value relevance function, which is able to detect that $[0.4, 0.6)$ is the most relevant interval.}
	\label{fig:shapley_relevance_fn_example}
\end{figure}

An illustrative example is shown in Figure \ref{fig:shapley_relevance_fn_example}. On the top left panel, there is a functional data set with 100 curves. The interval considered is $I = [0,1]$ and each function is generated as $\mathcal{X}(t) = (t - 0.5)^2 + \varepsilon(t)$ for $t$ in a fine grid of points, where $\varepsilon(t) \sim N(\mu=0, \sigma=0.01)$ and $\varepsilon(t_1)$ is independent of $\varepsilon(t_2)$ if $t_1 \neq t_2$. The target variable is generated as $Y = \Upsilon(\mathcal{X})= \min_{t \in I}[\mathcal{X}(t)]$. The top right panel of Figure \ref{fig:shapley_relevance_fn_example} shows the density of $\argmin[\mathcal{X}(t)]$.

Let us suppose that $I$ is divided into 5 parts (of the same length), that is, the players considered are $I_1, \ldots, I_5$. Since most of the density of $\argmin[\mathcal{X}(t)]$ is in the interval $[0.4, 0.6)$, when using a prediction model with a good predictive capacity, it is expected that the Shapley value relevance function assigns the majority of the relevance to player $I_3 = (0.4, 0.6]$. Let us use the true regression function as predictive function: $f(\mathcal{X})= \Upsilon(\mathcal{X})$. The bottom left and right panels of Figure \ref{fig:shapley_relevance_fn_example} depict respectively the histogram-type and the polygonal-type Shapley value relevance functions, which have the expected behavior.

\subsection{Estimating conditional expectations}
\label{sec:cond_expectation}

In this section we describe how $\dot{\mathcal{X}}_j(t) = \mathbb{E}\big( \boldsymbol{\mathcal{X}}(t)\mid \{\boldsymbol{\mathcal{X}}(u)=\mathcal{X}_j(u): u \in S\}\big)$ is estimated, $1 \leq j \leq m$. For this purpose, it is convenient to assume that $\boldsymbol{\mathcal{X}}$ is a Gaussian process observed at the set of points $\mathcal{T} = \{t_1, \ldots, t_T\}$. Then, $(\boldsymbol{\mathcal{X}}(t_1),\ldots,\boldsymbol{\mathcal{X}}(t_T))$ follows a multivariate normal distribution (with mean $\boldsymbol{\mu}$ and variance matrix $\boldsymbol{\Sigma}$) from which there are $m$ independent observations. Let $\mathbf{X}$ be the resulting $m\times T$ data matrix.

Let $S$ be the subset of $I$ which is assumed to be available (or observed). Let $\mathcal{T}_{S} = S \cap \mathcal{T}$ and let $\mathcal{J}_{S}$ be the corresponding indexes in $\{1, \ldots, T\}$. We refer to $\mathcal{J}^{\mathsf{c}}_{S}$ as the complementary set of $\mathcal{J}_{S}$. For the sake of clarity, we use the labels ``O" for the {\em observed} set $\mathcal{J}_{S}$ and ``M" for the {\em missing} set $\mathcal{J}^{\mathsf{c}}_{S}$. The estimated variance covariance matrix from $\mathbf{X}$, $\boldsymbol{\hat{\Sigma}}$, can be rearrange as
\begin{equation*}
    \boldsymbol{\hat{\Sigma}} =
    \begin{bmatrix}
        \boldsymbol{\hat{\Sigma}}_{MM} & \boldsymbol{\hat{\Sigma}}_{MO} \\
        \boldsymbol{\hat{\Sigma}}_{OM} & \boldsymbol{\hat{\Sigma}}_{OO}
    \end{bmatrix}.
\end{equation*}

Let $\hat{\boldsymbol{\mu}}_O$ (respectively, $\hat{\boldsymbol{\mu}}_M$) be the estimation of the mean vector for those columns whose indexes belong to $\mathcal{J}_{S}$ (respectively, $\mathcal{J}^{\mathsf{c}}_{S}$). Let $\mathbf{x}_{jO}$ be the vector that results from selecting the $j$-th row of $\mathbf{X}$ and those columns whose indexes belong to $\mathcal{J}_{S}$. Therefore, the standard theory of multivariate normal distributions states that
\begin{gather}
    \{\hat{\dot{\mathcal{X}}}_j(t_h):h\in \mathcal{J}^{\mathsf{c}}_{S} \} = \hat{\mathbb{E}}\big( \{\boldsymbol{\mathcal{X}}(t_h): h\in \mathcal{J}^{\mathsf{c}}_{S} \}\mid \{\boldsymbol{\mathcal{X}}(t_d)=\mathcal{X}_j(t_d): d \in \mathcal{J}_{S}\}\big) = \notag \\
    \label{eq:CondExpectNormality}
    \hat{\boldsymbol{\mu}}_M + \boldsymbol{\hat{\Sigma}}_{MO} \boldsymbol{\hat{\Sigma}}_{OO}^{-1}(\mathbf{x}_{jO} - \hat{\boldsymbol{\mu}}_O).
\end{gather}

For the cases where $\boldsymbol{\hat{\Sigma}}_{OO}$ is not invertible, instead of $\boldsymbol{\hat{\Sigma}}_{OO}^{-1}$ we use $\boldsymbol{\hat{\Sigma}}_{OO}^{+}$, the Moore--Penrose pseudo-inverse of $\boldsymbol{\hat{\Sigma}}_{OO}$ computed by diagonalizing it, replacing all non-zero eigenvalues with their inverses, and leaving the zero eigenvalues as they are. It is worth remembering that Equation (\ref{eq:CondExpectNormality}) gives also the expression of the least squares linear prediction of $\boldsymbol{\mathcal{X}}_j(t_h)$, $h\in \mathcal{J}^{\mathsf{c}}_{S}$, given the values $\mathcal{X}_j(t_d)$, $d \in \mathcal{J}_{S}$, even if multivariate normality is not assumed.

In terms of execution time, it is desirable to provide an expression to compute $\{\hat{\dot{\mathcal{X}}}_j(t_h):h\in \mathcal{J}^{\mathsf{c}}_{S} \}$ for all $1 \leq j \leq m$ simultaneously. The matrix-wise expression is given by
\begin{equation*}
    \mathbf{1}_m \hat{\boldsymbol{\mu}}_M^\intercal + \left(\mathbf{X}_O - \mathbf{1}_m \hat{\boldsymbol{\mu}}_O^\intercal\right) \boldsymbol{\hat{\Sigma}}_{OO}^{-1} \boldsymbol{\hat{\Sigma}}_{OM},
\end{equation*}
where $\mathbf{1}_m $ is a vector of length $m$ full of 1's and $\mathbf{X}_O$ is the matrix derived by selecting those columns whose indexes belong to $\mathcal{J}_{S}$ from $\mathbf{X}$.

\subsection{Reconstructing partially observed functional data}

The definition of a complete function $\tilde{\mathcal{X}}_j^S$ from $\mathcal{X}_j$, when only the values $\mathcal{X}_j(u)$ for $u\in S$ are known, is closely related to the problem of reconstructing partially observed functional data, which has received certain attention in the FDA literature. See, for instance, \citet{Goldberg2014,Kraus2015,Kneip2020}. In this last paper, the authors give the theoretical expression of the optimal linear reconstruction operator, which minimizes the pointwise mean squared prediction error for $\mathcal{X}_j(u)$, for any $u \in S^{\mathsf{c}}$:
\begin{equation}
    \label{eq:optimal_reconstruction}
	\hat{\mathcal{X}}_j(u)=\mu(u) + \sum_{k=1}^\infty \xi_{jk}^O \frac{\langle \gamma_u, \phi_k^O \rangle}{\lambda_k^O},
\end{equation}
where $\mu(t)=\mathbb{E}(\boldsymbol{\mathcal{X}}(t))$, $t\in I$, is assumed to be known, $(\lambda_k^O,\phi_k^O)$, $k\ge 1$, are the pairs of non-zero eigenvalues and eigenfunctions of the covariance operator of the functional data restricted to the observed set $S$ (corresponding to the covariance function $\gamma^O(t,s)=\mbox{Cov}(\boldsymbol{\mathcal{X}}(t),\boldsymbol{\mathcal{X}}(s))$, $t$ and $s$ in $S$), $\langle\cdot,\cdot\rangle$ denotes the inner product in $L^2(S)$, $\mu^O$ and $\mathcal{X}_j^O$ are the restrictions of $\mu$ and $\mathcal{X}_j$ on $S$, respectively, $\xi_{jk}^O = \langle \phi_k^O, \mathcal{X}_j^O -\mu^O\rangle$ is the score of $\mathcal{X}_j^O$ in the $k$-th principal function computed from $\gamma^O(\cdot,\cdot)$, and $\gamma_u(t)= \mbox{Cov}(\boldsymbol{\mathcal{X}}(u),\boldsymbol{\mathcal{X}}(t))$, $t$ in $S$.

Let us show that the completion formula given by Equation (\ref{eq:CondExpectNormality}) is a feasible version of the optimal reconstruction operator (\ref{eq:optimal_reconstruction}). We call $(\hat{\lambda}_k^O, \hat{\boldsymbol{\phi}}_k^O)$, $k=1,\ldots,r_O$, the pairs of non-zero eigenvalues and eigenvectors of the sampling covariance matrix $\boldsymbol{\hat{\Sigma}}_{OO}$. Then
\begin{equation*}
    \boldsymbol{\hat{\Sigma}}_{OO}= \sum_{k=1}^{r_O} \hat{\lambda}_k^O \hat{\boldsymbol{\phi}}_k^O \left(\hat{\boldsymbol{\phi}}_k^O\right)^T,
    \text{ and }
    \boldsymbol{\hat{\Sigma}}_{OO}^+= \sum_{k=1}^{r_O} \frac{1}{\hat{\lambda}_k^O} \hat{\boldsymbol{\phi}}_k^O \left(\hat{\boldsymbol{\phi}}_k^O\right)^T.
\end{equation*}
Let $\hat{\xi}_{jk}^O=\left(\hat{\boldsymbol{\phi}}_k^O\right)^T (\mathbf{x}_{jO} - \hat{\boldsymbol{\mu}}_O)$ be the score of $\mathbf{x}_{jO}$ in the $k$-th principal component computed from $\boldsymbol{\hat{\Sigma}}_{OO}$. Consider an index $h\in \mathcal{J}^{\mathsf{c}}_S$ and let $u=t_h$. Let $\hat{\boldsymbol{\gamma}}_u^T$ be the corresponding row of $\boldsymbol{\hat{\Sigma}}_{MO}$, which contains estimations of $\mbox{Cov}(\boldsymbol{\mathcal{X}}(u),\boldsymbol{\mathcal{X}}(t_d))$, $d$ in $\mathcal{J}_S$. The row of the expression in Equation (\ref{eq:CondExpectNormality}) corresponding to $u=t_h$ is
\begin{equation*}
    \hat{\dot{\mathcal{X}}}_j(u)
    = \hat{\mu}(u) + \hat{\boldsymbol{\gamma}}_u^T \sum_{k=1}^{r_O} \frac{1}{\hat{\lambda}_k^O} \hat{\boldsymbol{\phi}}_k^O \left(\hat{\boldsymbol{\phi}}_k^O\right)^T (\mathbf{x}_{jO} - \hat{\boldsymbol{\mu}}_O)
    = \hat{\mu}(u) + \sum_{k=1}^{r_O} \hat{\xi}_{jk}^O \frac{\hat{\boldsymbol{\gamma}}_u^T \hat{\boldsymbol{\phi}}_k^O}{\hat{\lambda}_k^O},
\end{equation*}
which coincides with Equation (\ref{eq:optimal_reconstruction}), once the unknown elements are estimated.

\subsection{Functional relevance is not based on a vector measure game}
\label{sec:No_vector_measure_game}

In this section we show that, even in quite simple situations, the continuous games used for defining functional relevance are not vector measure games. Therefore, we can not expect to find a generic closed form expression (as the diagonal formula mentioned in Section \ref{sec:continuous_games}) for the relevance function. Assume that the observed data are $(\mathcal{X}_j,y_j)$, $j\in\{1,\ldots,m\}$, where the responses $y_j$ are generated by a linear functional model: $Y=\int_0^1 \beta(t) \boldsymbol{\mathcal{X}}(t) dt + \varepsilon$, with $\mathbb{E}(\varepsilon)=0$ and $\mbox{Var}(\varepsilon)=\sigma^2$. We fit a linear scalar-on-function regression model to estimate the functional coefficient $\beta(t)$. Assume that the estimation is so precise that we can operate as if $\beta(t)$ is known.

Consider the payoff function $\nu(S)$ defined in Equation (\ref{eq:payoff_R2}), which defines the continuous game used to find the relevance function for the fitted scalar-on-function regression model. Observe that $\nu(S)$ depends on $S$ only through
{\allowdisplaybreaks
\begin{align*}
	\frac{1}{m} \sum_{j=1}^m (y_j - \tilde{y}_j^S)^2 
	& = \frac{1}{m} \sum_{j=1}^m \left(\int_0^1 \beta(t) \mathcal{X}_j(t) dt +\varepsilon_j - \int_0^1 \beta(t) \tilde{\mathcal{X}}_j^S(t) dt \right)^2 \\
	& = \frac{1}{m} \sum_{j=1}^m \left( \int_{S^{\mathsf{c}}} \beta(u)\left( \mathcal{X}_j(u) - \dot{\mathcal{X}}_j^S(u) \right) du +\varepsilon_j \right)^2 \\
	& \approx \text{Var}\left(\int_{S^{\mathsf{c}}} \beta(u)\left( \boldsymbol{\mathcal{X}}(u) - \boldsymbol{\dot{\mathcal{X}}}^S(u) \right) du +\varepsilon\right) \\
	& = \text{Var}\left(\int_{S^{\mathsf{c}}} \beta(u) \boldsymbol{\mathcal{E}}^S(u) du\right) + \sigma^2 \\
	& = \int_{S^{\mathsf{c}}} \int_{S^{\mathsf{c}}} c_{\mathcal{E}}^S(u,v) \beta(u)\beta(v) du dv + \sigma^2 \\
	& = \int_{S^{\mathsf{c}}} \int_{S^{\mathsf{c}}} \sum_{k=1}^\infty \lambda_k^S \xi_k^S(u) \xi_k^S(v) \beta(u)\beta(v) du dv + \sigma^2 \\
	& = \sum_{k=1}^\infty \lambda_k^S \left( \int_{S^{\mathsf{c}}} \xi_k^S(u) \beta(u) du \right)^2 + \sigma^2,
\end{align*}
}%
where $\boldsymbol{\mathcal{E}}^S=\boldsymbol{\mathcal{X}} - \boldsymbol{\dot{\mathcal{X}}}^S$ is the reconstruction error (which is assumed to have zero mean; see \citealp[Theorem 2.2]{Kneip2020}), $c_{\mathcal{E}}^S(u,v)$ is the covariance function of $\boldsymbol{\mathcal{E}}^S$, and $(\lambda_k^S, \xi_k)$, $k\ge 1$, are the eigenvalues and eigenfunctions of the covariance operator of $\boldsymbol{\mathcal{E}}^S$.

Let $\tau_k^S$, $k\ge 1$, be signed measures defined as $\tau_k^S(B)=\int_{B} \xi_k^S(u) \beta(u) du$ for any Borel set $B\subseteq I$, and let $\tau_k^{S+}$ and $\tau_k^{S-}$ be the positive and negative parts of $\tau_k$. We have shown that $\nu(S)$ is approximately equal to a function of
\begin{equation*}
    \sum_{k=1}^\infty \lambda_k^S \tau_k^S(S^{\mathsf{c}})^2 = \sum_{k=1}^\infty \lambda_k^S \left(\tau_k^S(I) - \tau_k^S(S) \right)^2.
\end{equation*}
Therefore $\nu(S)\approx\rho_S(\{\tau_k^{S+}(S),\tau_k^{S-}(S):k\ge 1\}))$ for a function $\rho_S$ defined on $\mathbb{R}^{\mathbb{N}}$. We conclude that, in general, the payoff function $\nu$ used to define functional relevance does not correspond to a vector measure game because: (i) the function $\rho^S$ depends on a countably infinite set of measure functions, and (ii) $\rho^S$, as well as the measures $\tau_k^{S+}$ and $\tau_k^{S-}$, depend on the set $S$ on which $\nu$ is evaluated.

\section{The \pkg{ShapleyFDA} package}
\label{sec:package}

An open source \proglang{Python} package, \pkg{ShapleyFDA}, has been released to PyPI to compute the Shapley value relevance function according to the methodology explained in this work. Given that the calculations are inherently matrix-intensive, the fundamental component of \pkg{ShapleyFDA} is based on the \pkg{NumPy} package \citep{Harris2020}.

As a large volume of computations must be considered, our implementation stores the intermediate results in memory for future reference. For instance, given permutations $\pi_1 = (I_3, I_4, I_1, I_2)$ and $\pi_2 = (I_3, I_4, I_2, I_1)$ the set of players preceding player 1 in $\pi_1$ is the same as the set of players preceding 2 in $\pi_2$. Upon the initial computation of the value for $p_{\pi_1}^1=\{I_3,I_4\}$, it is stored. This approach guarantees that the same value will not be calculated twice, but rather used as many times as necessary.

\pkg{ShapleyFDA} assumes that the prediction model has already been trained. Furthermore, the package is designed so that it can handle multiple prediction models simultaneously to obtain the Shapley value relevance function for each of them. Consequently, the same set of random permutations is used when obtaining multiple Shapley value relevance functions. Moreover, the input test data set must be provided in matrix format. The package incorporates functionalities to display both figures, the histogram-type function and the polygonal-type function. Apart from the released version in PyPI, there is also a development version available at GitHub (\url{https://github.com/pachoning/ShapleyFDA}).

Now let us derive the complexity of our proposed algorithm. We consider Equation (\ref{eq:shapley_cont_formula}), where a random subset of permutations $\Pi_0 \subset \Pi$ is employed. First, we obtain the cost for a fixed permutation $\pi \in \Pi_0$, and then we consider the whole set of permutations. In order to compute $\tilde{R}^2(p_{\pi}^i \cup I_i)$, or $\tilde{R}^2(p_{\pi}^i)$, it is needed to first obtain $\tilde{\mathcal{X}}^S_j(t)$, as stated in Equation (\ref{eq:x_recumputed}). All the details on how to compute $\tilde{\mathcal{X}}^S_j(t)$ are provided in Section \ref{sec:cond_expectation}. Of particular interest is the matrix-wise expression
\begin{equation}
	\label{eq:computational_cost}
	\mathbf{1}_m \hat{\boldsymbol{\mu}}_M^\intercal + \left(\mathbf{X}_O - \mathbf{1}_m \hat{\boldsymbol{\mu}}_O^\intercal\right) \boldsymbol{\hat{\Sigma}}_{OO}^{-1} \boldsymbol{\hat{\Sigma}}_{OM},
\end{equation}
where subindex ``O'' refers to the $T_O$ {\em observed} columns of $\mathbf{X}$, and subindex ``M'' refers to the remaining $T_M$ {\em missing} columns. The computational complexity of Equation (\ref{eq:computational_cost}) is
\begin{equation*}
    \mathcal{O}\left(m T_O\right) + \mathcal{O}\left(m T_M\right) + \mathcal{O}\left(T_O^3\right) + \mathcal{O}\left(m T_O^2\right) + \mathcal{O}\left(m T_M T_O\right)
    =
    \mathcal{O}\left(T_O^3\right) + \mathcal{O}\left(m T_O^2\right) + \mathcal{O}\left(m T_M T_O\right),
\end{equation*}
where we use that the computational complexity of matrix multiplication of two matrices of sizes $s_1\times s_2$ and $s_2 \times s_3$ is $\mathcal{O}(s_1 s_2 s_3)$, and that the computational complexity of matrix inversion of a matrix of size $s_4 \times s_4$ is $\mathcal{O}(s_4^3)$ (see, for instance, \citealp[page 12]{Farebrother1988}). Observe that $T_O$ and $T_M$ are random variables (they depend on the random permutation $\pi$) and that $\mathbb{E}(T_O)=\mathbb{E}(T_M)=T/2$. Then, the expected computational complexity of Equation (\ref{eq:computational_cost}) is $\mathcal{O}\left(T^3\right) + \mathcal{O}\left(m T^2\right)$.

Let $\mathcal{O}_{f}(m,T)$ be the complexity of running the prediction algorithm for a matrix of size $m \times T$. Considering the $n$ players and the $\lvert \Pi_0 \rvert$ permutations, the global expected complexity of our proposal is $\mathcal{O}\left( T^3 \lvert \Pi_0 \rvert n\right) + \mathcal{O}\left( m T^2 \lvert \Pi_0 \rvert n\right) + \mathcal{O}_{f}(m,T)\mathcal{O}\left( \lvert \Pi_0 \rvert n \right)$.

\section{Experiments}
\label{sec:experiments}

Two types of experiments are conducted with the objective of showing the performance of our proposed methodology. In the first type, described in Section \ref{sec:exp_simulated_data}, we simulate data and then use three different prediction models to predict the response variable. Since we control the data generation process, we know in advance which points $t\in I$ are important. The second type, which is explained in Section \ref{sec:exp_real_data}, consists of exploring the Tecator data set \citep{Borggaard1992}. The code utilized in the experimental configuration is accessible via GitHub (\url{github.com/pachoning/shapley_fda_experiments}). Simulated data are also available at \url{https://bit.ly/4crYVt1}.

\subsection{Simulated data}
\label{sec:exp_simulated_data}

\begin{figure}
	\centering
	\begin{subfigure}[t]{.32\textwidth}
		\centering
		\includegraphics[width=\textwidth]{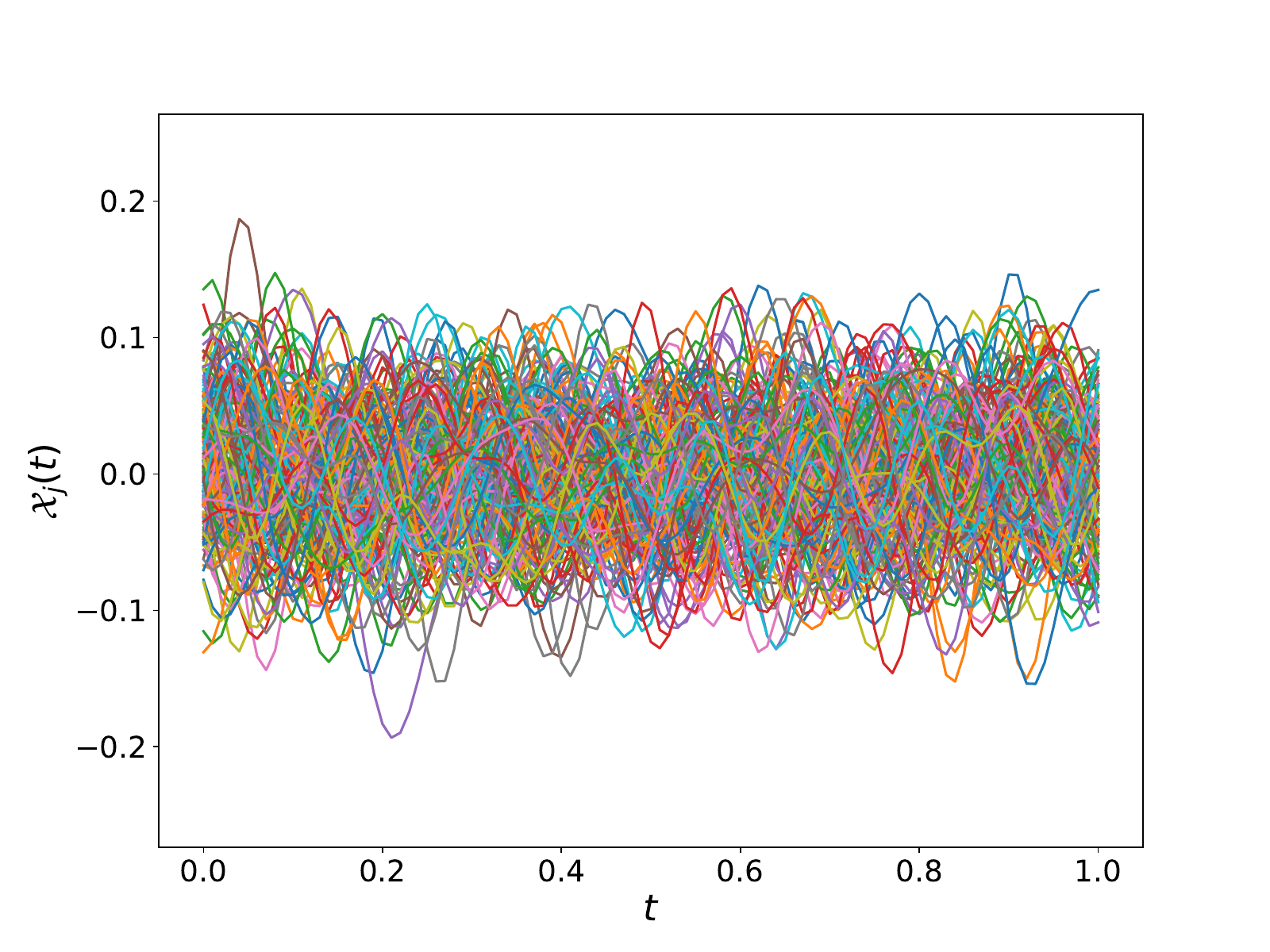}
		\caption{Fourier expansion.}
		\label{fig:functional_covariates_fourier}
	\end{subfigure}
	\hfill
	\begin{subfigure}[t]{.32\textwidth}
		\centering
		\includegraphics[width=\textwidth]{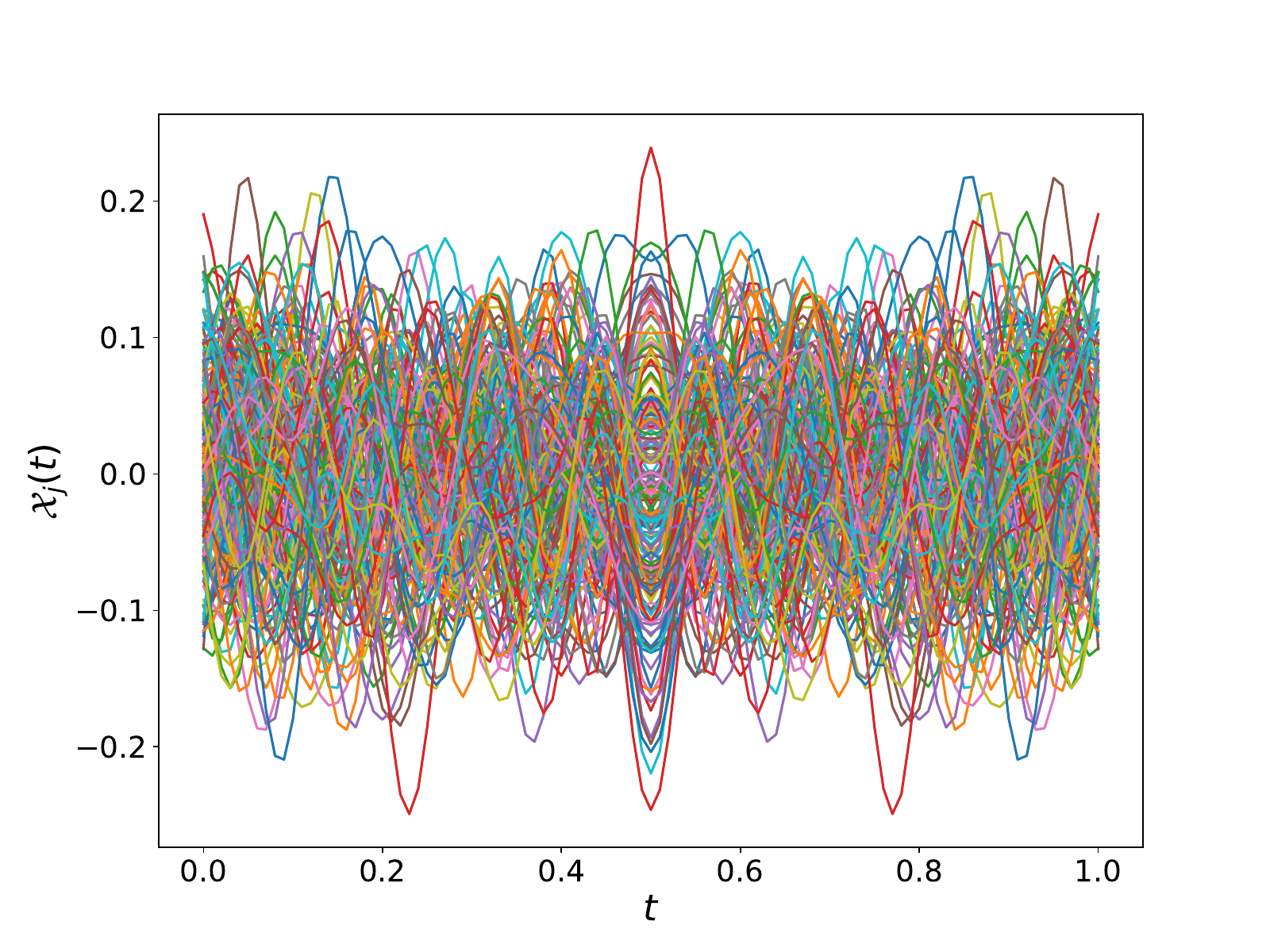}
		\caption{Symmetric Fourier expansion.}
		\label{fig:functional_covariates_symmetric}
	\end{subfigure}
	\hfill
	\begin{subfigure}[t]{.32\textwidth}
		\centering
		\includegraphics[width=\textwidth]{all_functions_brownian.pdf}
		\caption{Brownian trend.}
		\label{fig:functional_covariates_brownian}
	\end{subfigure}
	
	\caption{Functional data sets.}
	\label{fig:functional_covariates}
\end{figure}

Regarding $\boldsymbol{\mathcal{X}}$, we consider three different ways of generation, namely {\em Fourier expansion}, {\em symmetric Fourier expansion} and {\em Brownian trend} (short name for {\em Brownian motion with a trend}). The first one consists of linear combinations of the Fourier basis, $\boldsymbol{\mathcal{X}}_{\mathrm{F}}(t) = \sum_{s=0}^r Z_s\xi_s(t)$, where $Z_s$ are independent draws of a standard normal distribution, while $\{\xi_s(t)\}_{s\ge 0}$ is the Fourier basis on $I$, and $r$ is an even number. The goal of the second one is to obtain symmetric trajectories with respect to the middle point of $I$. This is achieved by building the functions as $\boldsymbol{\mathcal{X}}_{\mathrm{sF}}(t) = \boldsymbol{\mathcal{X}}_{\mathrm{F}}(t) + 	\boldsymbol{\mathcal{X}}_{\mathrm{F}}(1-t)$. When simulating data for the previous two cases, $\boldsymbol{\mathcal{X}}_{\mathrm{F}}(t)$ and $\boldsymbol{\mathcal{X}}_{\mathrm{sF}}(t)$, the value of $r$ is fixed at 30. The third functional random variable taken into account is a Brownian motion with a trend. Let $\boldsymbol{\mathcal{B}}(t)$ be a Brownian motion for $t \in I=[0,1]$. Then, the functional random variable is $\boldsymbol{\mathcal{X}}_{\mathrm{B}}(t) = \boldsymbol{\mathcal{B}}(t) + t$. Figure \ref{fig:functional_covariates} illustrates the aforementioned functional random variables.

\begin{figure}
	\centering
	\begin{subfigure}[t]{.49\textwidth}
		\centering
		\includegraphics[width=\textwidth]{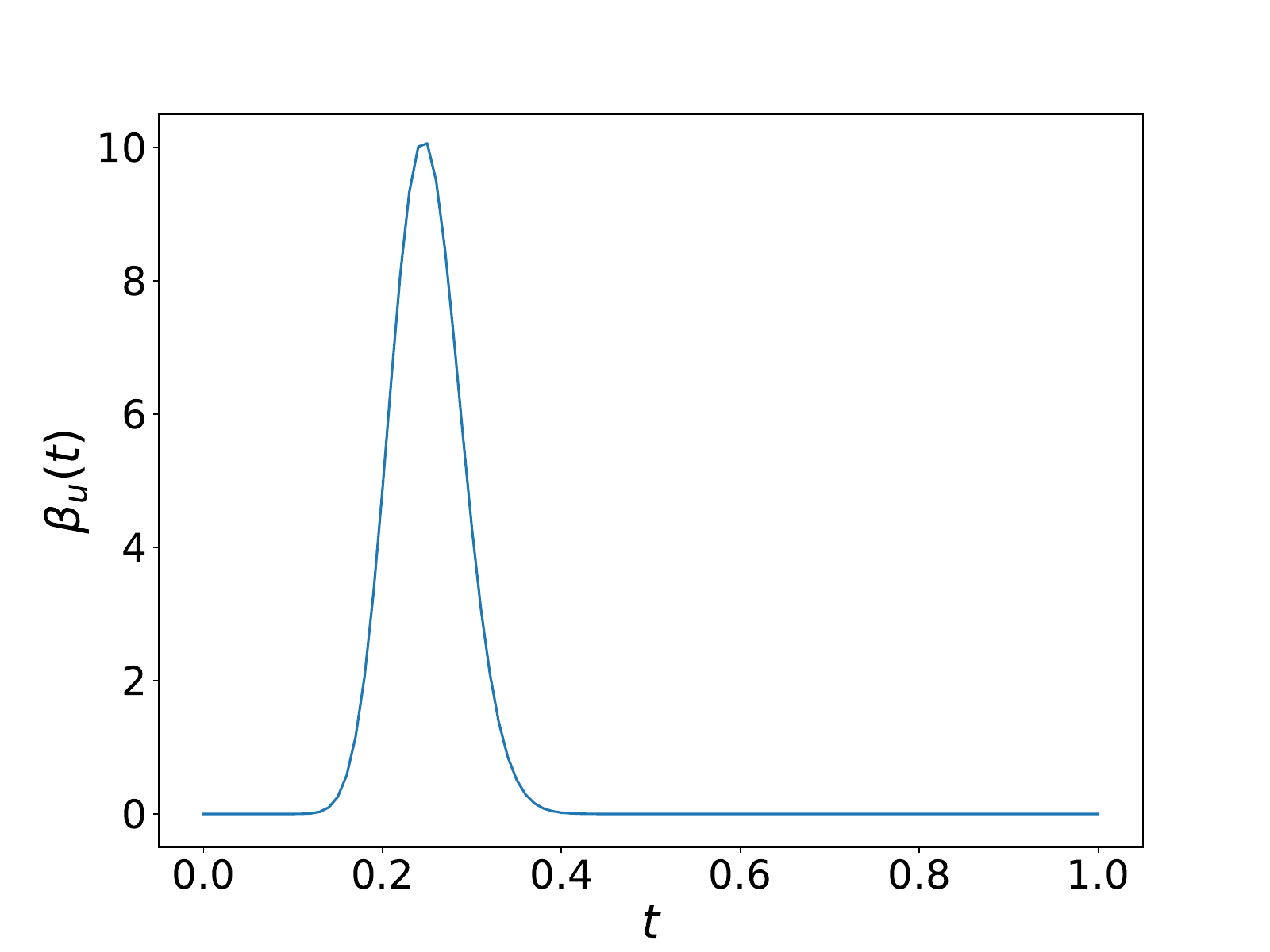}
		\caption{$\mathrm{Beta}(30, 90)$}
		\label{fig:betas_distributions_beta}
	\end{subfigure}
	\hfill
	\begin{subfigure}[t]{.49\textwidth}
		\centering
		\includegraphics[width=\textwidth]{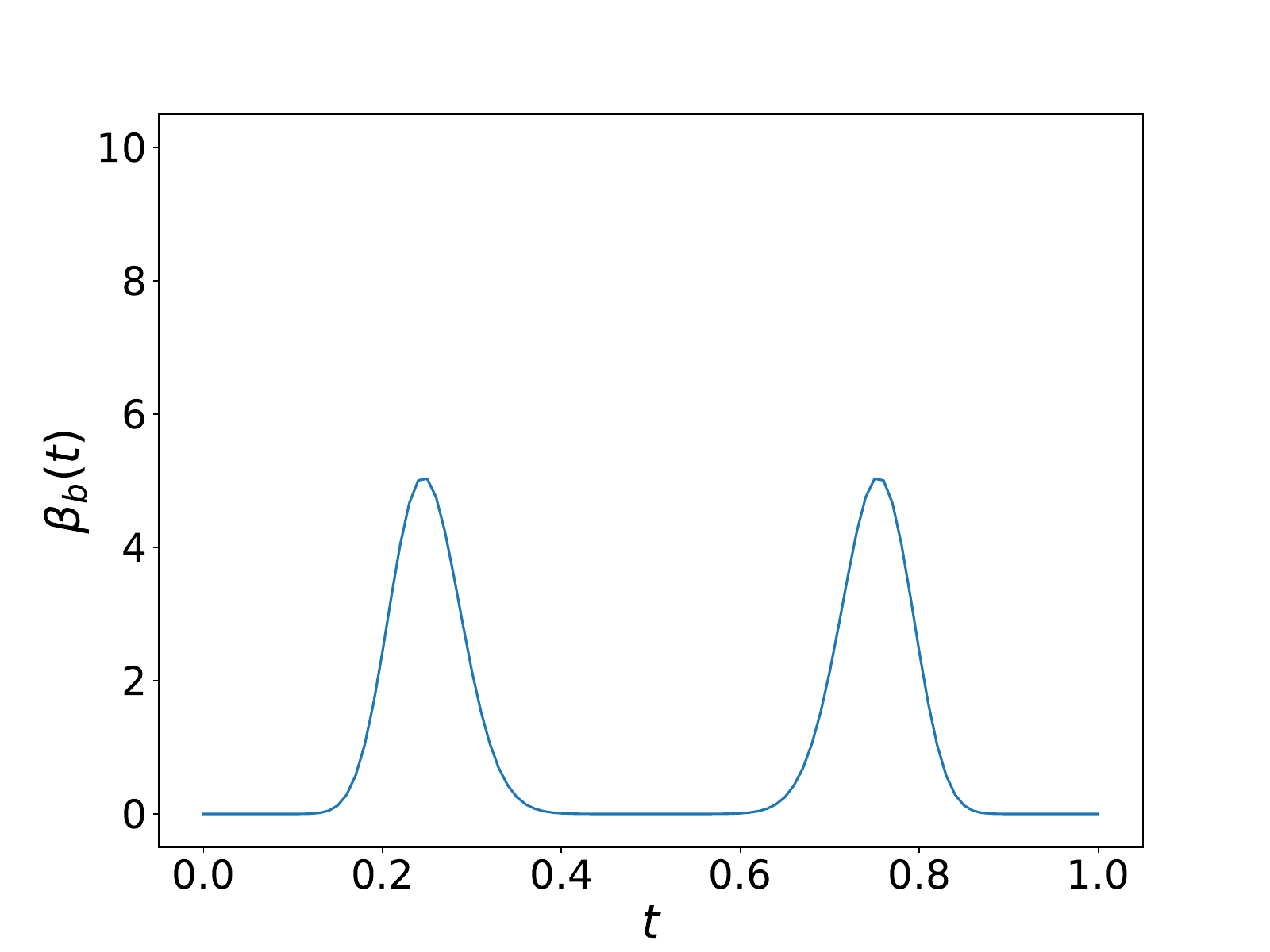}
		\caption{$\frac{1}{2}\mathrm{Beta}(30, 90) + \frac{1}{2}\mathrm{Beta}(90, 30)$}
		\label{fig:betas_distributions_mixture}
	\end{subfigure}
	
	\caption{Beta density functions.}
	\label{fig:betas_distributions}
\end{figure}

Every functional data set is stored in a matrix $\mathbf{X}$ of size $m \times T$, where $m$ takes two different values, 200 and 500, and $T$ is equal to $101$, being $t_1 = 0$ and $t_k = t_{k-1} + 0.01$, $k=2,\ldots, 101$. Regarding the partition $\mathcal{I} = \{I_1, \ldots, I_n\}$, we consider $n=20$ and all the intervals with the same length. In order to build the target variable $Y$, a transformation is applied to the functional random variable, $\Upsilon: L^2(I) \rightarrow \mathbb{R}$, leading to $Y = \Upsilon\left(\boldsymbol{\mathcal{X}}\right) + \epsilon$, being $\epsilon \sim N(0, \sigma_{\epsilon}^2)$ independent of $\boldsymbol{\mathcal{X}}$. We control the signal-to-noise ratio by using $\eta=\sigma_{\epsilon}^2/\sigma_{Y}^2$, where $\sigma_{Y}^2 = \text{Var}(Y)$. We use two values for $\eta$: 0.05 and 0.25. With respect to $\Upsilon$, 4 types of transformations are taken into account:
\begin{description}
	\item[Linear unimodal:] $\Upsilon_{\mathrm{lu}}\left(\boldsymbol{\mathcal{X}}\right) = \int_I \boldsymbol{\mathcal{X}}(t) \beta_{\mathrm{u}}(t) dt$, with $\beta_{\mathrm{u}}(t)$ being the density function of a beta distribution with parameters $30$ and $90$. See left panel of Figure \ref{fig:betas_distributions}.
	\item[Linear bimodal:] $\Upsilon_{\mathrm{lb}}\left(\boldsymbol{\mathcal{X}}\right) = \int_I \boldsymbol{\boldsymbol{\mathcal{X}}}(t) \beta_{\mathrm{b}}(t) dt$, with $\beta_{\mathrm{b}}(t) = (1/2)[\beta_{\mathrm{u}}(t) + \beta_{\mathrm{u}}(1-t)]$. See right panel of Figure \ref{fig:betas_distributions}.
	\item[Non-linear:] $\Upsilon_{\mathrm{nl}}\left(\boldsymbol{\mathcal{X}}\right) = \max_{t \in I} \{\lvert  \beta_{\mathrm{u}}(t) \boldsymbol{\mathcal{X}}(t) \rvert, \lvert \beta_{\mathrm{u}}(t) \boldsymbol{\mathcal{X}}(1 - t) \rvert \}$.
	\item[Discrete:] $\Upsilon_{\mathrm{d}}\left(\boldsymbol{\mathcal{X}}\right) = \boldsymbol{\mathcal{X}}(0.15) + \lvert \boldsymbol{\mathcal{X}}(0.55) \rvert + \boldsymbol{\mathcal{X}}^2(0.35)\boldsymbol{\mathcal{X}}(0.85)$.
\end{description}
In total, there are 48 different scenarios, the result of combining all the possibilities:
\begin{equation*}
    \{\boldsymbol{\mathcal{X}}_{\mathrm{F}}(t),\boldsymbol{\mathcal{X}}_{\mathrm{sF}}(t),\boldsymbol{\mathcal{X}}_{\mathrm{B}}(t)\} \times \underbrace{\{200, 500\}}_m \times \underbrace{\{0.05, 0.25\}}_\eta \times \{\Upsilon_{\mathrm{lu}}, \Upsilon_{\mathrm{lb}}, \Upsilon_{\mathrm{nl}}, \Upsilon_{\mathrm{d}}\}.
\end{equation*}

The relationship between $\boldsymbol{\mathcal{X}}$ and $Y$ is modeled using three distinct prediction algorithms: a functional linear regression model (FLM for short; \citealp[Chapter 12]{Ramsay2005}), a functional k-nearest neighbor algorithm (FKNN for short; \citealp[Chapter 7]{Ferraty2006}) and a functional version of neural networks (FNN for short; \citealp{Heinrichs2023}). As for the software used to train those models, \pkg{scikit-fda} \citep{RamosCarreno2024} contains an implementation of FLM and FKNN. In addition, Florian Heinrichs has an open source implementation of FNN available at GitHub (\url{github.com/FlorianHeinrichs/functional_neural_networks}).

Every scenario is replicated 100 times. A replica is constituted by the generation of three independent data sets: training, validation and test. The first one is used to train the prediction models. As those algorithms require hyperparameter tuning, the second data set is used to do it. Finally, the goal of the third data set is to compute the Shapley value relevance function for each prediction algorithm. These three sets are of the same shape, $m \times T$. So, given a scenario, there are 100 sets of Shapley value relevance functions. A total of $\lvert \Pi_0 \rvert = 1000$ random permutations of the set $\mathcal{I}$ are generated.

In the context of basis representation, a preliminary analysis showed that the number of elements in the basis is more important than the type of basis itself. To this end, we decided to use splines as the basis for FLM and FKNN, and the default Legendre basis for FNN. The number of elements constitutes an hyperparameter for each of the models.

Shapley value relevance functions are obtained for each prediction algorithm and each scenario, as well as for each of the 100 simulations. These results are averaged across simulations. For the sake of brevity, we offer here the results corresponding to sample size $200$ and signal-to-noise ratio given by $\eta=0.05$ (12 scenarios out of 48; the remaining results are reported as Appendix \ref{sec:results_simulation_study} in the supplementary material). Figure \ref{fig:shapley_scenarios_m_200_eta_005} shows the mean Shapley value relevance functions, and Table \ref{table:r2_m_200_eta_005} contains the mean and the standard deviation (in brackets) across the 100 simulations of $R^2(I)$, that is the coefficient of determination using all points $t\in [0, 1]$ evaluated on the test data sets.

\begin{figure}
	\centering
	
	\generategridfigures{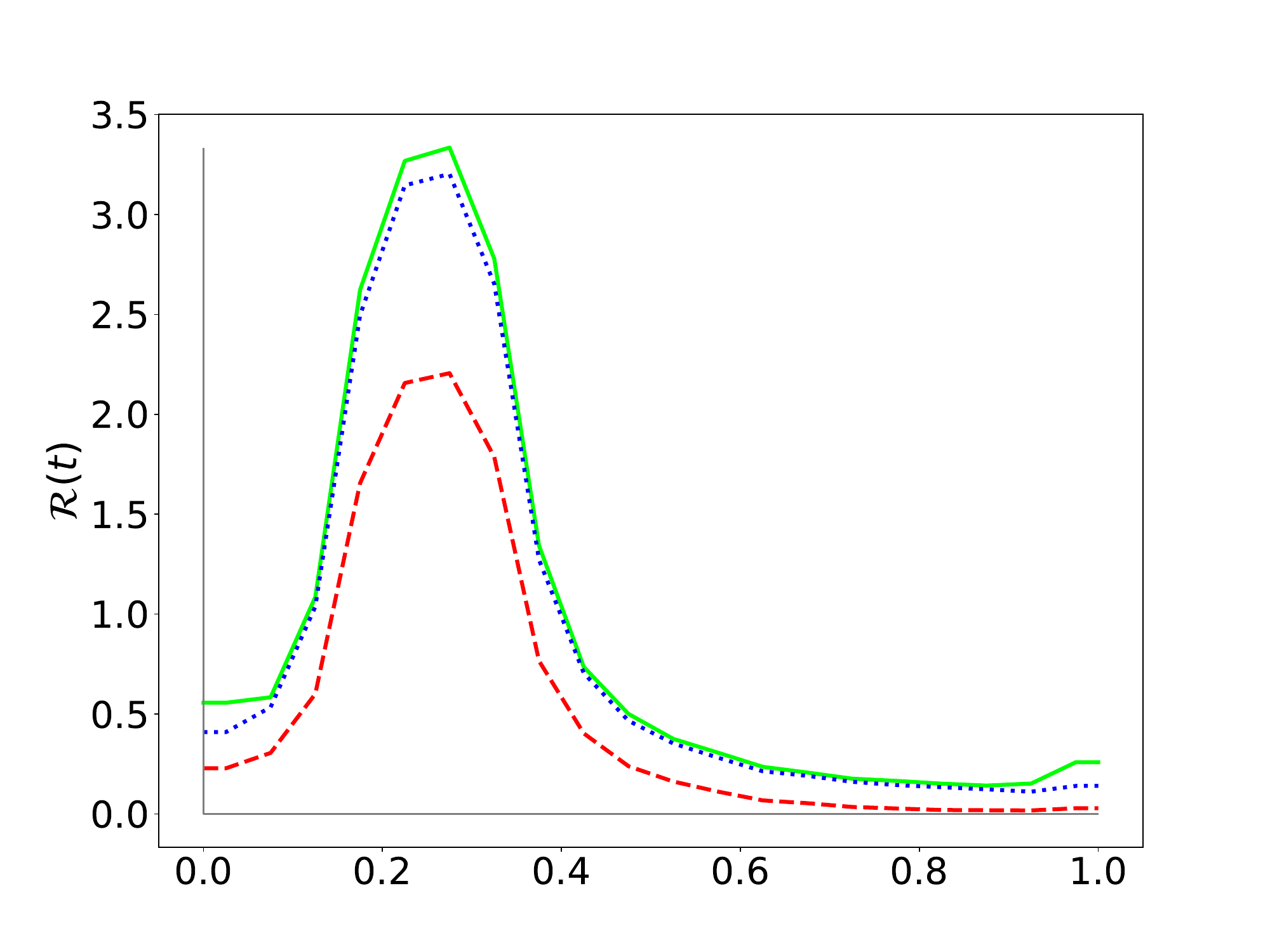,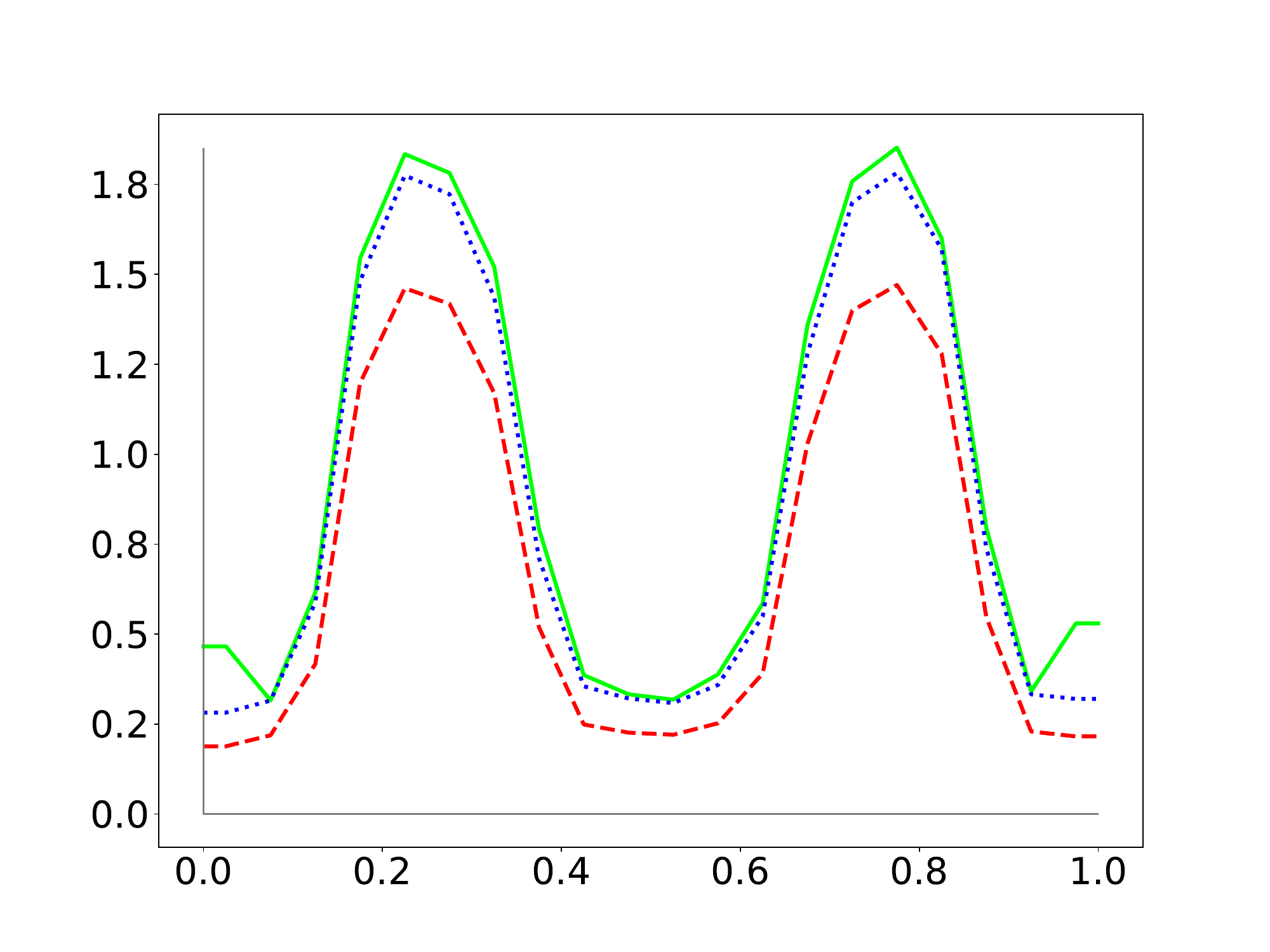,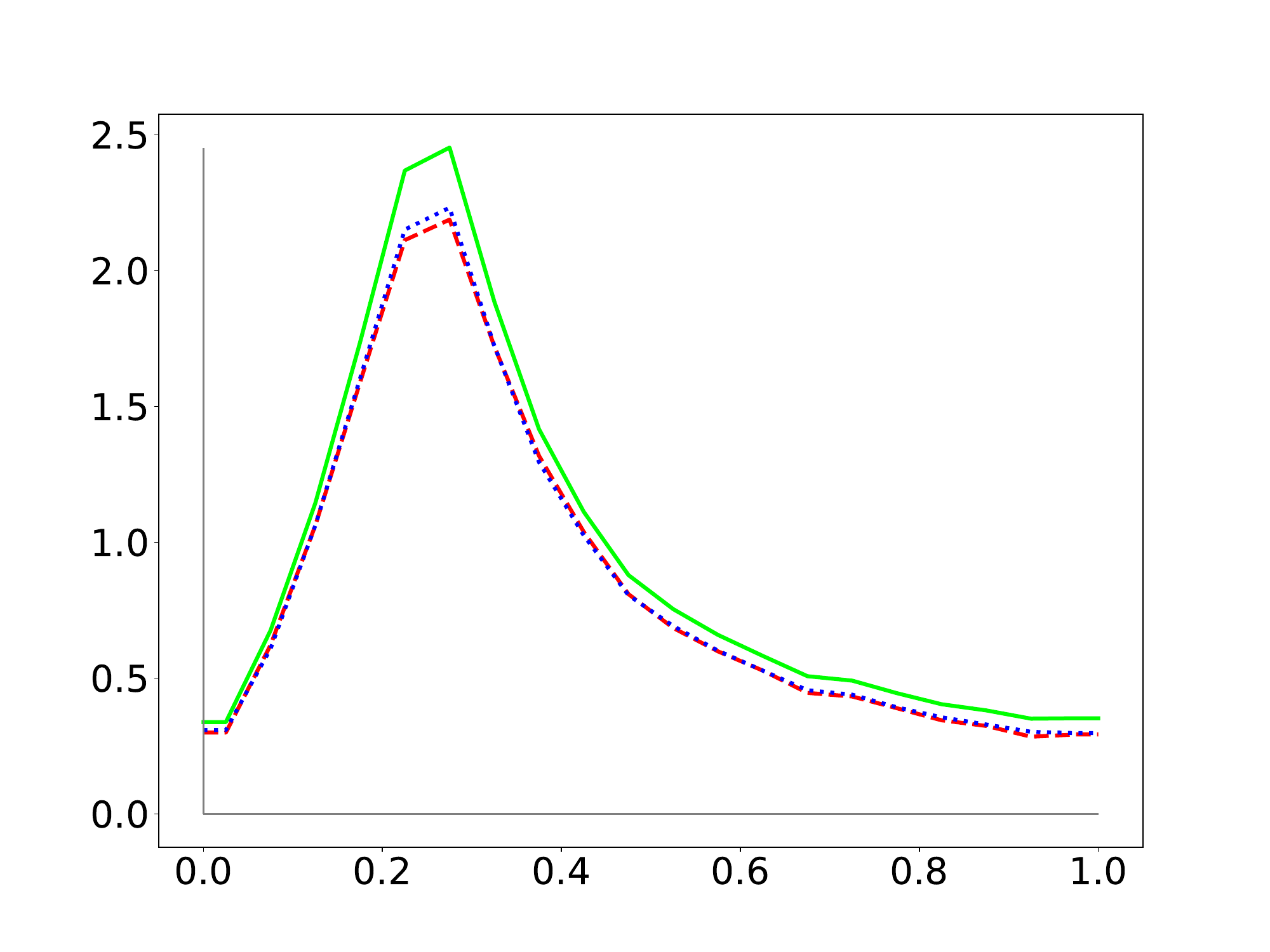,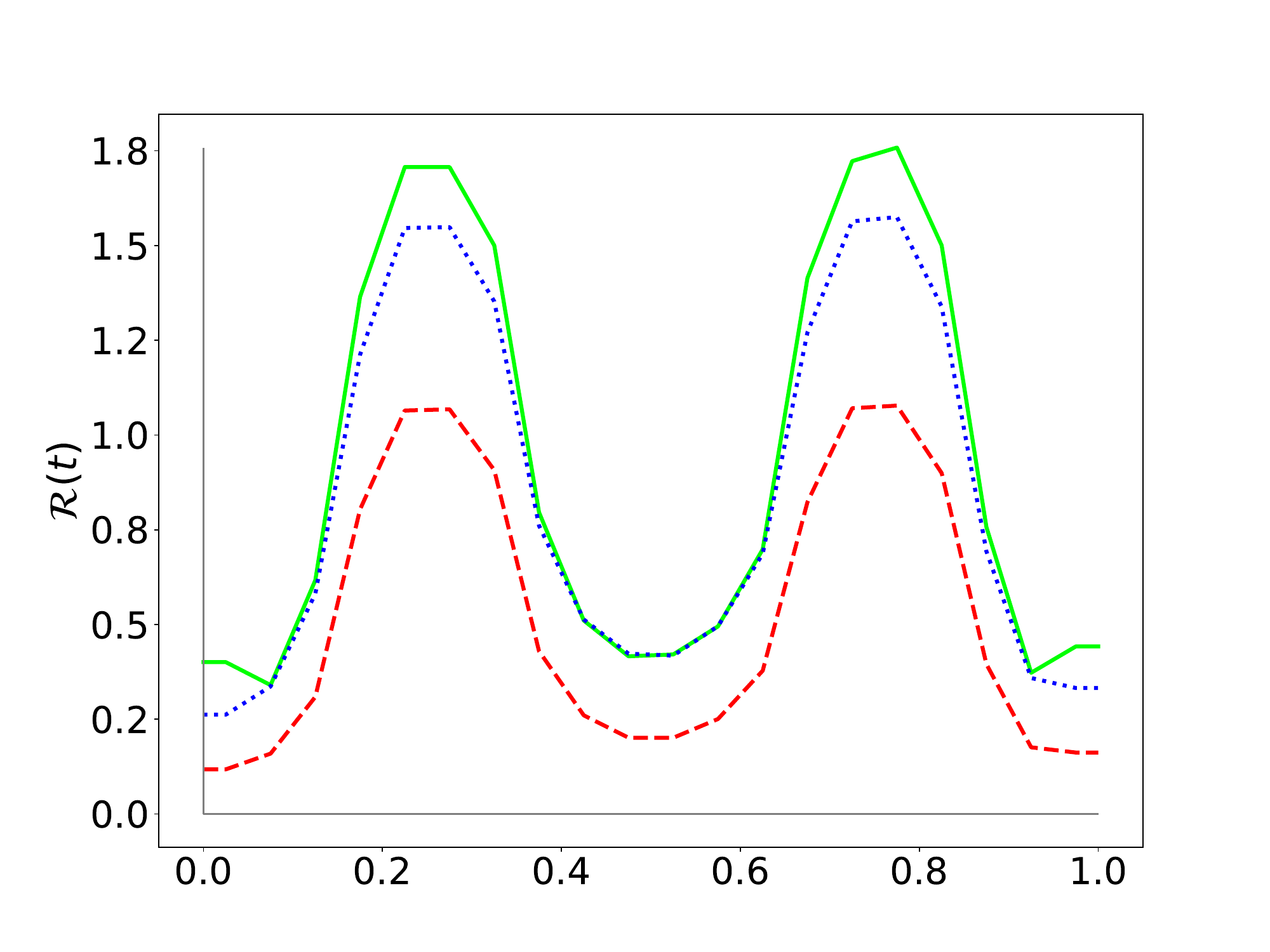,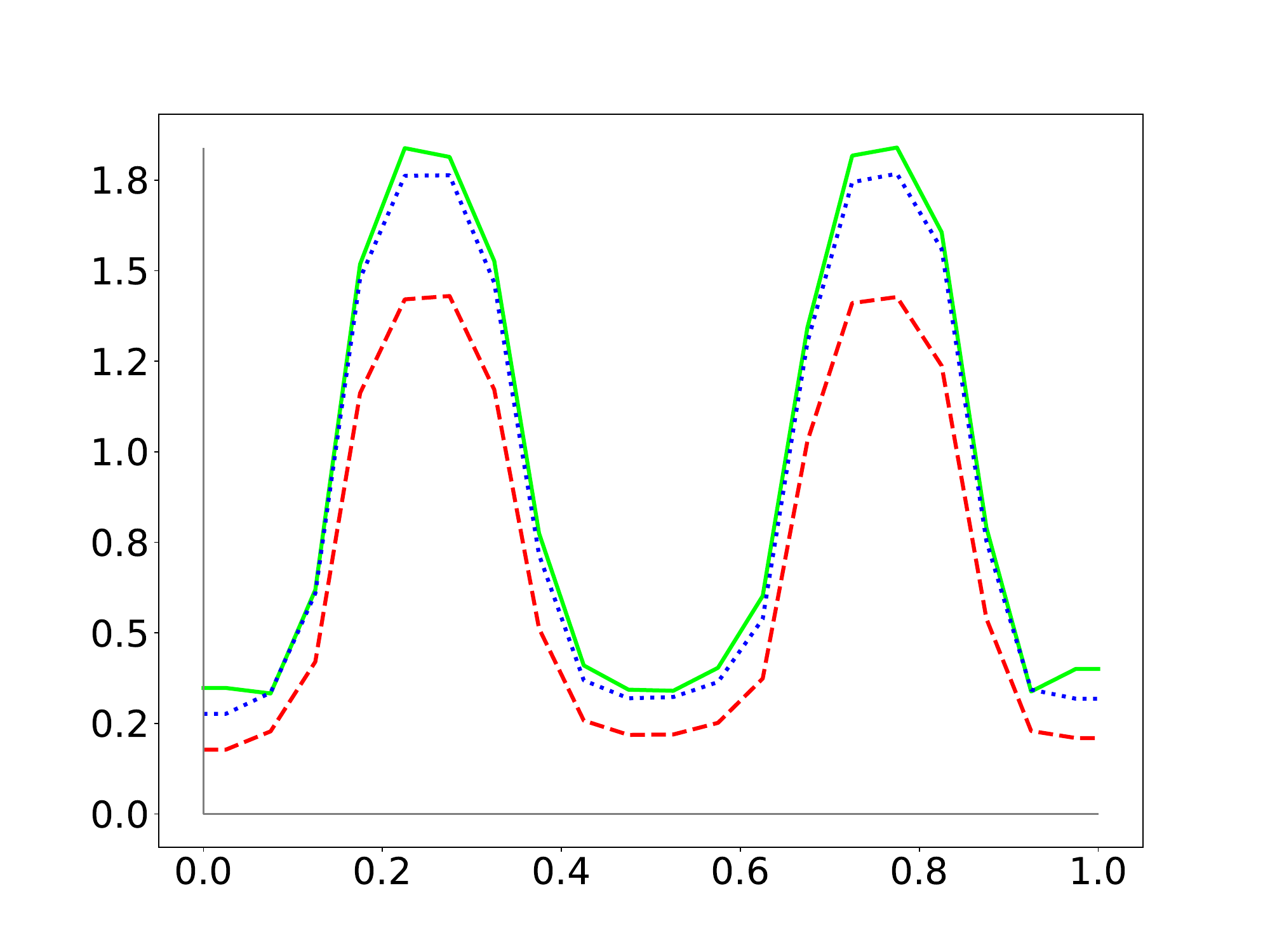,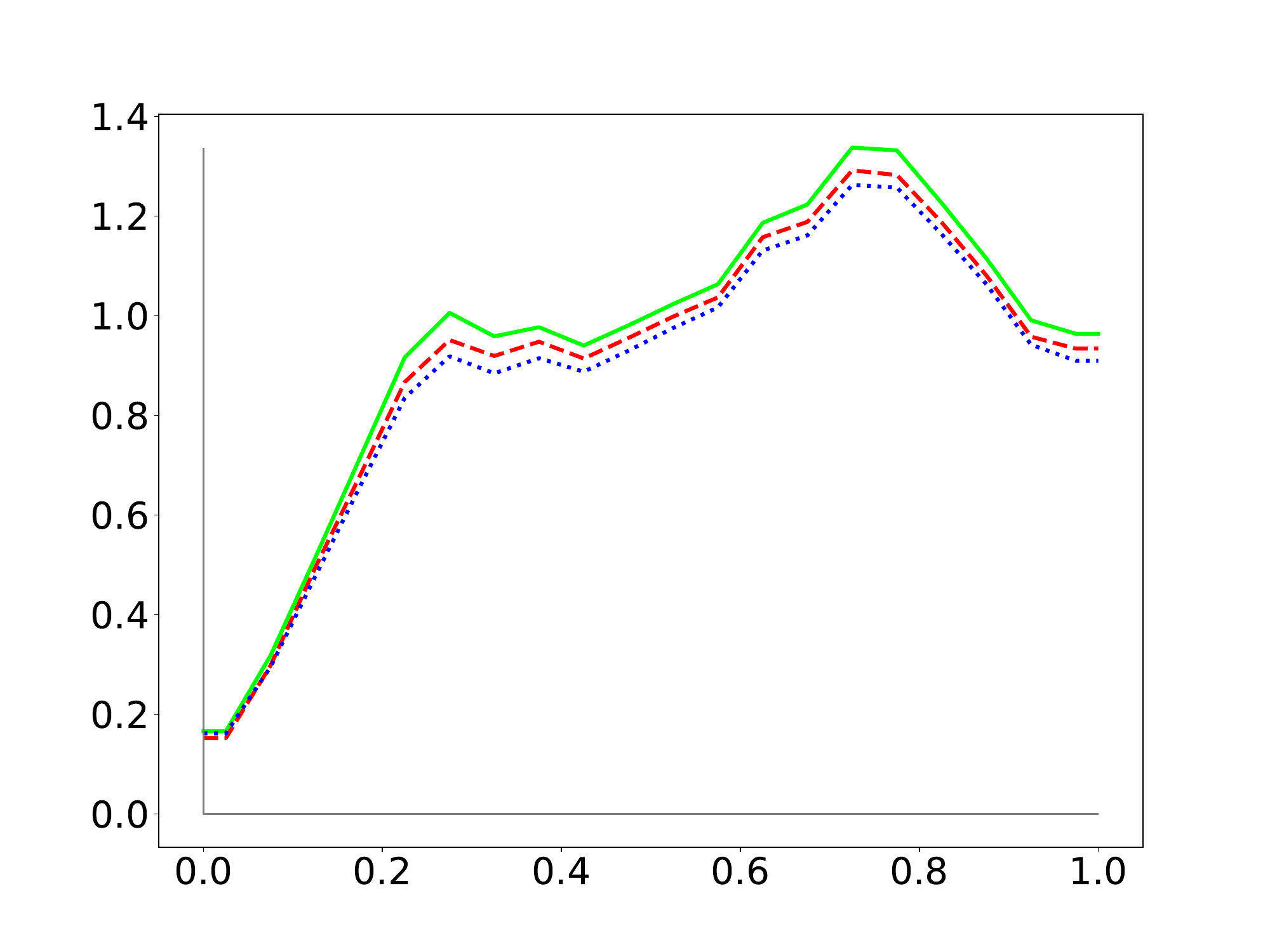,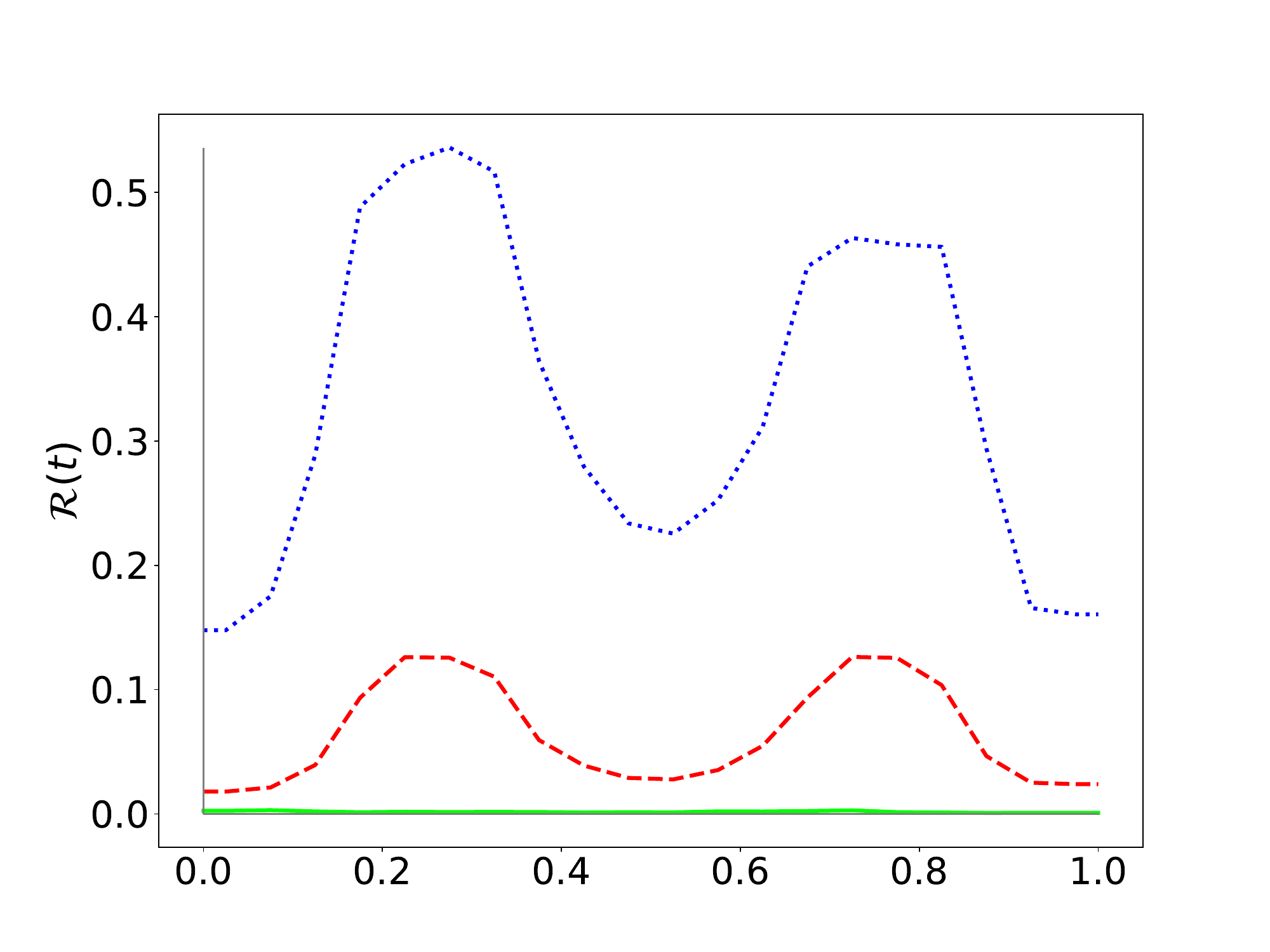,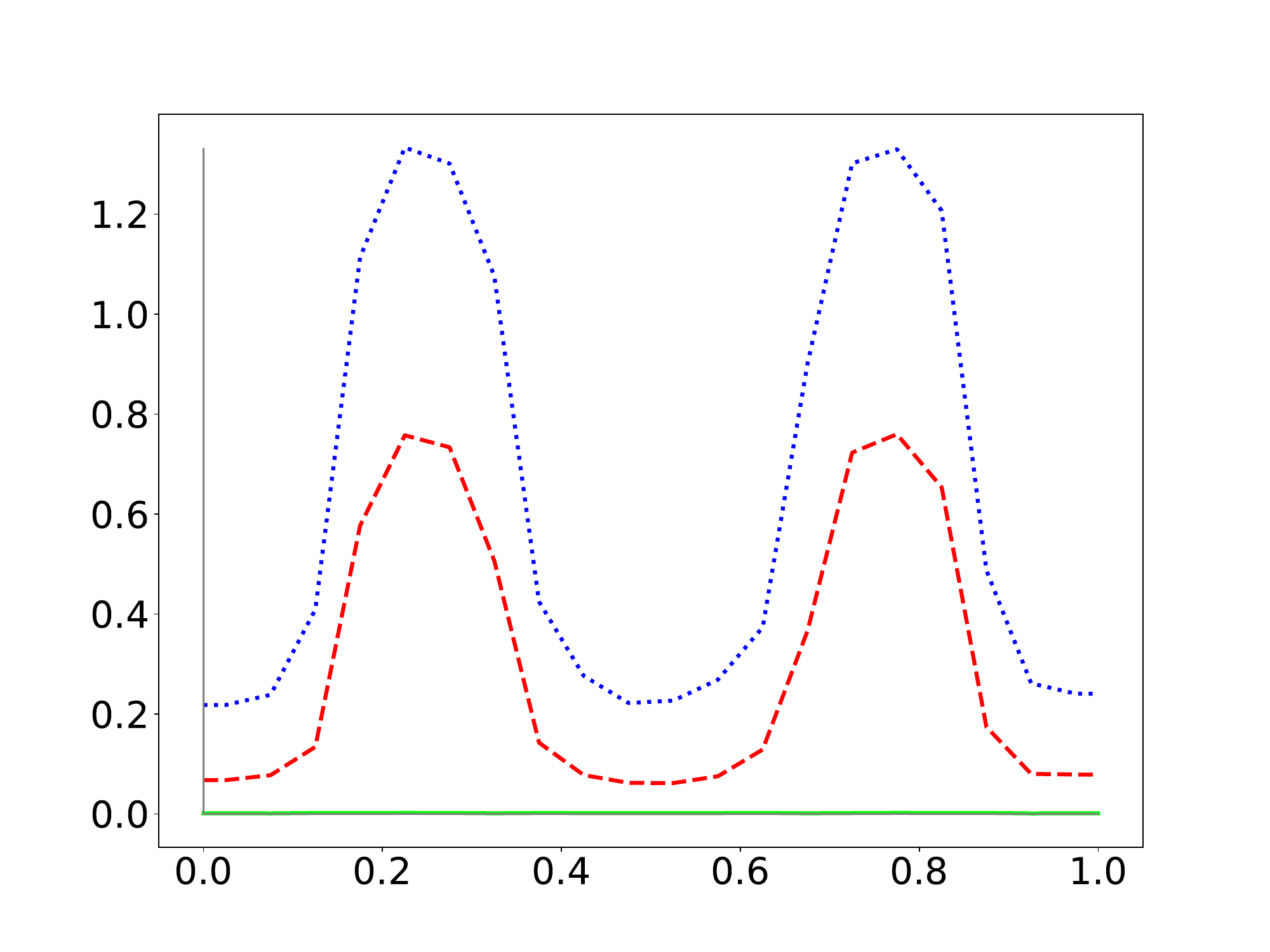,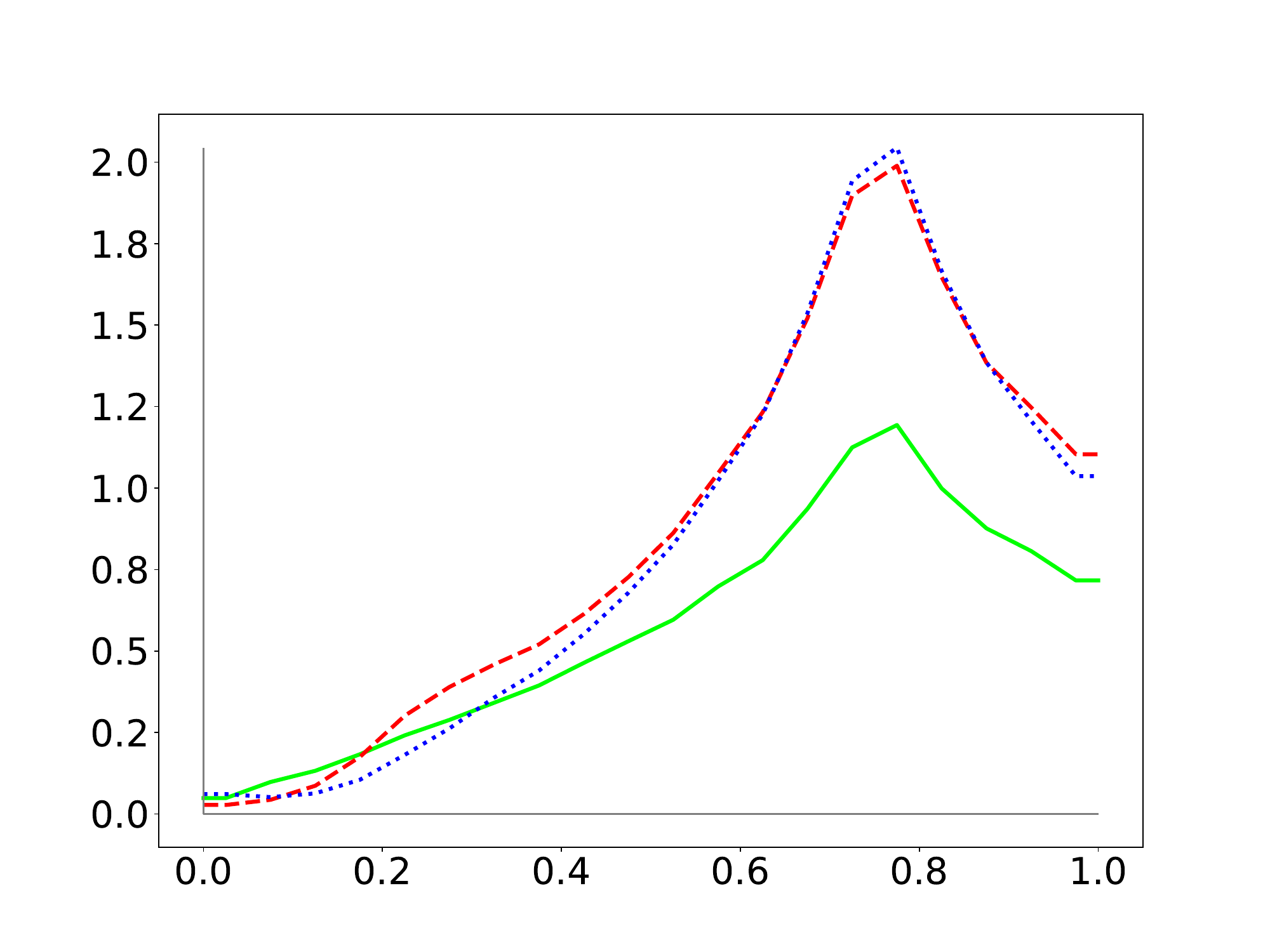,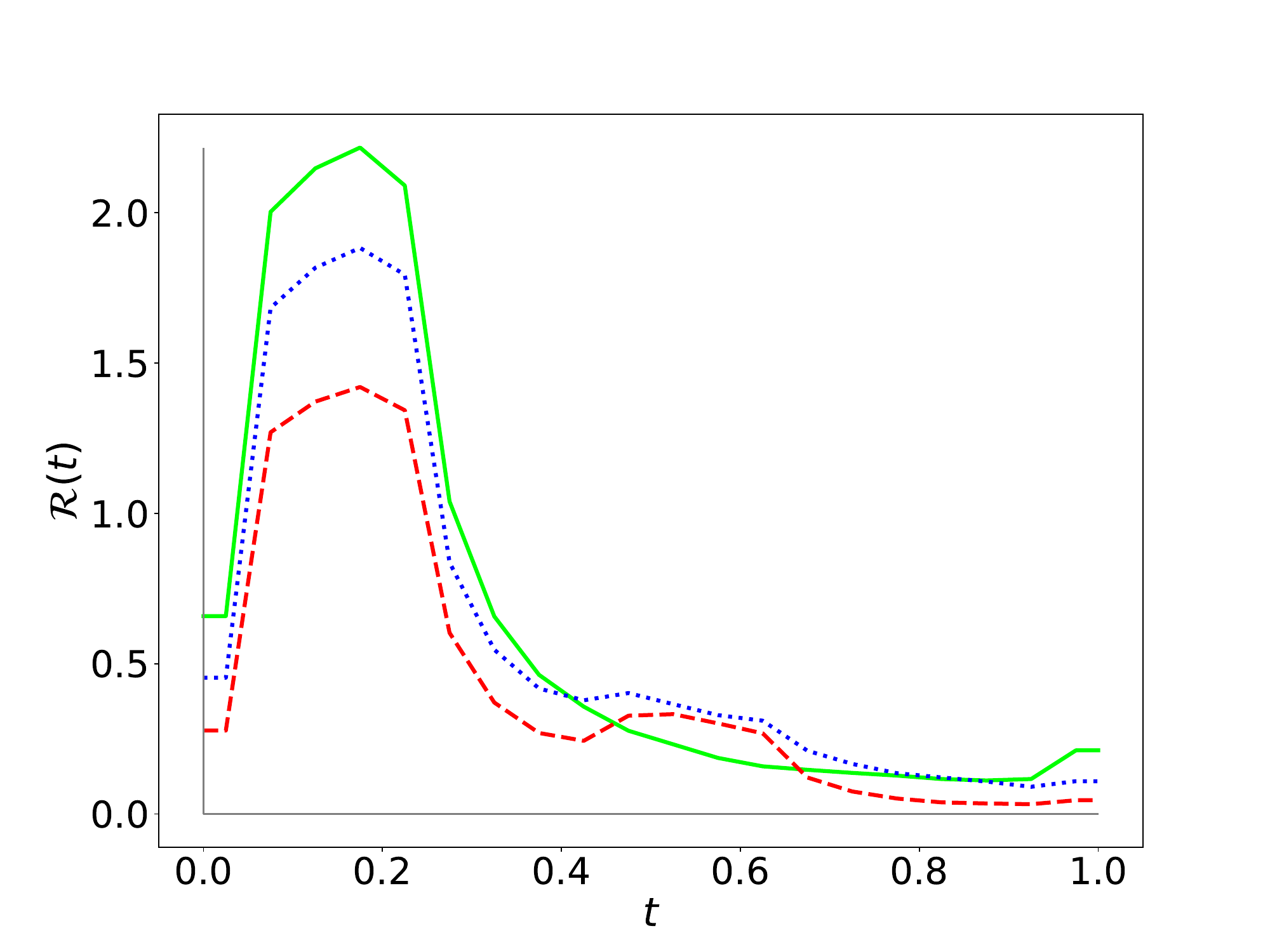,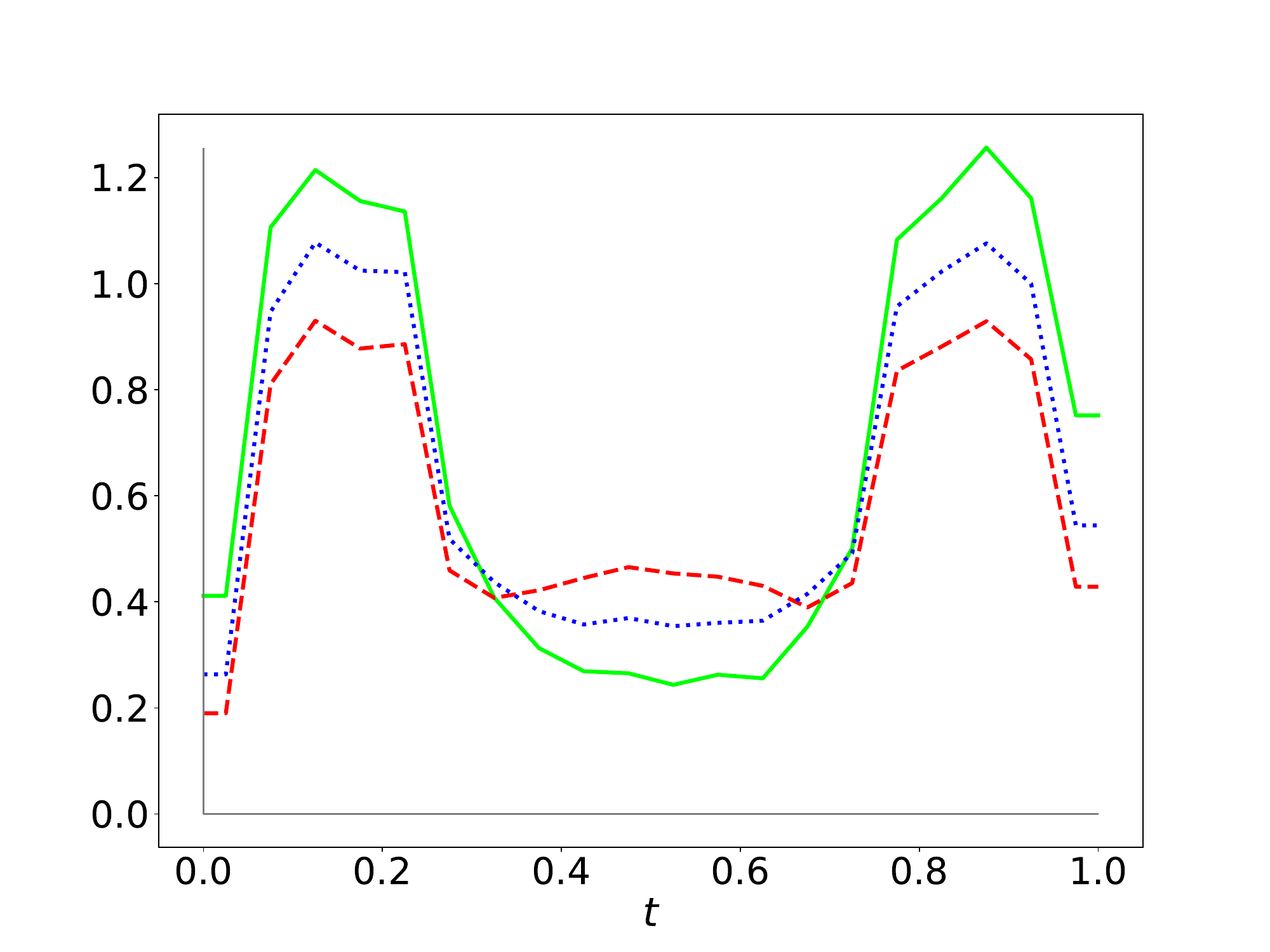,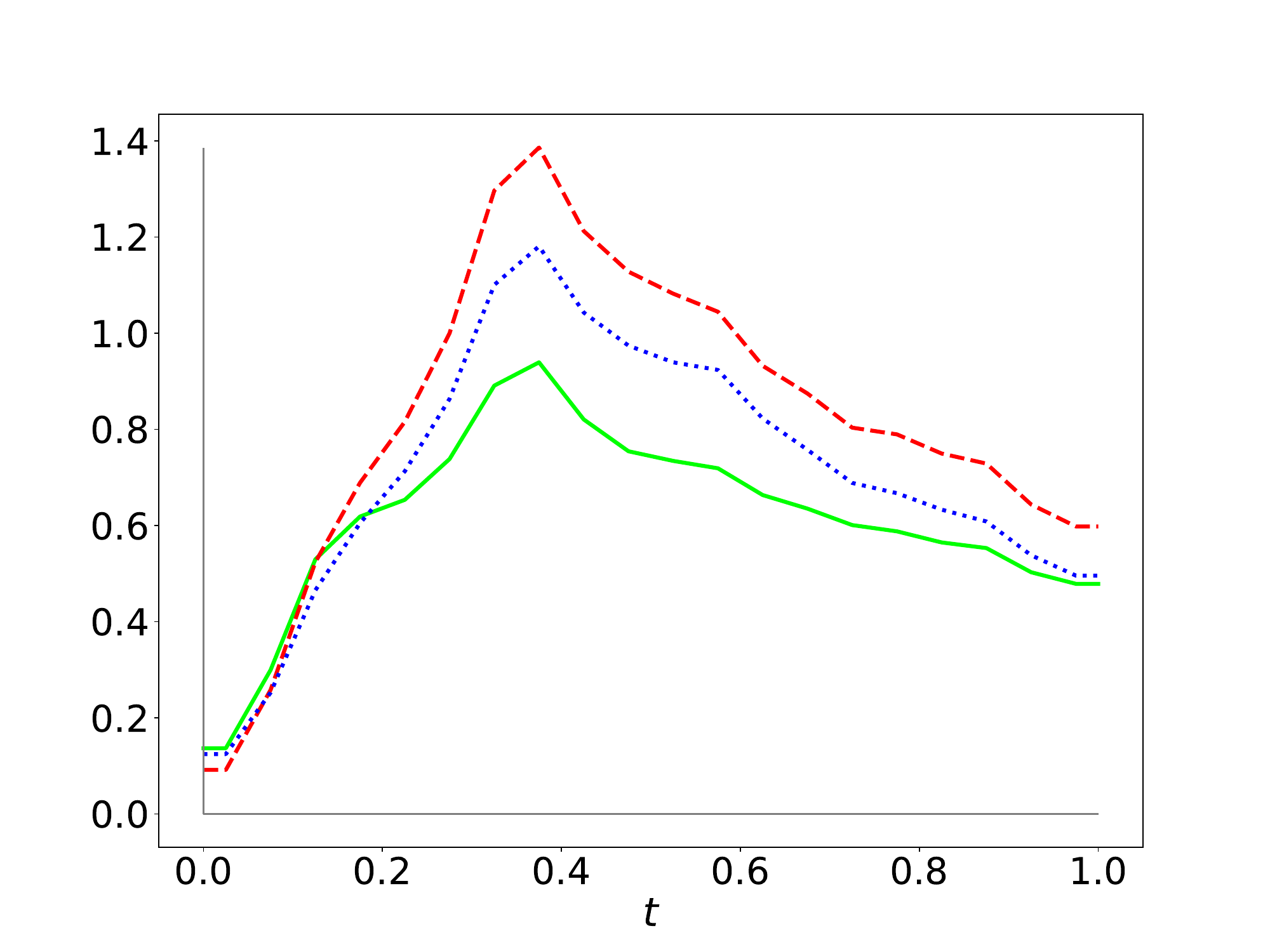}
	
	\includegraphics[width=.4\textwidth]{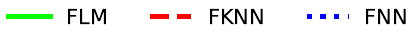}
	\caption{Mean Shapley value relevance functions for those scenarios with $m=200$ and $\eta=0.05$.}
	\label{fig:shapley_scenarios_m_200_eta_005}
\end{figure}

\begin{table}
	\centering

	\generategridtable
	{0.9468 (0.0054),0.9483 (0.0053),0.9467 (0.0058),0.5399 (0.0392),0.7048 (0.0259),0.8541 (0.0191),0.8887 (0.0812),0.8961 (0.0902),0.8602 (0.1160)}
	{0.9464 (0.0062),0.9470 (0.0053),0.9479 (0.0051),0.5377 (0.0370),0.6962 (0.0273),0.9149 (0.0091),0.8634 (0.0726),0.9053 (0.0719),0.8924 (0.0521)}
	{0.0289 (0.0331),0.0264 (0.0326),0.5718 (0.0753),0.0490 (0.0804),0.3101 (0.0668),0.8637 (0.0163),0.3361 (0.1450),0.6559 (0.1602),0.8321 (0.0721)}
	{0.6696 (0.0388),0.6907 (0.0383),0.6209 (0.0462),0.4337 (0.0397),0.5990 (0.0372),0.8324 (0.0464),0.6070 (0.1568),0.6492 (0.0909),0.7196 (0.0644)}

	\caption{Mean value (standard deviation) of $R^2(I)$ of FLM, FKNN and FNN for those scenarios with $m=200$ and $\eta=0.05$.}
	\label{table:r2_m_200_eta_005}
\end{table}

First, we consider the scenarios with functional data generated as Fourier expansions (first column in Figure \ref{fig:shapley_scenarios_m_200_eta_005}). When the target variable is derived from the linear unimodal or bimodal transformations (first and second row respectively), all the Shapley value relevance functions are able to identify the area with the highest relevance, which is the one formed by the points where the functional coefficients ($\beta_{\mathrm{u}}(t)$ or $\beta_{\mathrm{b}}(t)$) take their maximum values. Regarding the coefficients of determination $R^2(I)$, they indicate good fitting for both, FLM and FNN, and a poor performance of FKNN.

The third scenario (first column, third row in Figure \ref{fig:shapley_scenarios_m_200_eta_005}), with target of non-linear type, shows that the Shapley value relevance functions may be different from each other. First, it can be seen that the Shapley value relevance function corresponding to the functional linear model is constantly equal to 0. According to it, no point is relevant for the model. As the relationship between $\boldsymbol{\mathcal{X}}$ and $Y$ is non-linear, the estimated FLM is of very poor quality. This can be corroborated with the value of its $R^2(I)$, which mean value is close to 0 (see Table \ref{table:r2_m_200_eta_005}). Therefore, it is expected not to find any relevant point $t \in I$ when using the FLM. However, FKNN and FNN models are able to detect the relevant points in prediction, with FNN showing a better performance than FKNN.

Consider now the fourth set of Shapley value relevance functions in the first column of Figure \ref{fig:shapley_scenarios_m_200_eta_005}. It corresponds to using a discrete transformation when defining the target variable, $\Upsilon_{\mathrm{d}}\left(\boldsymbol{\mathcal{X}}\right) = \boldsymbol{\mathcal{X}}(0.15) + \lvert \boldsymbol{\mathcal{X}}(0.55) \rvert + \boldsymbol{\mathcal{X}}^2(0.35)\boldsymbol{\mathcal{X}}(0.85)$, which depends only on the values of $\mathcal{X}$ at four points ($0.15$, $0.35$, $0.55$ and $0.85$). The three Shapley value relevance functions indicate that the most relevant points $t$ are those around $0.15$, and that small to no relevance is assigned to the points $0.35$, $0.55$ or $0.85$. Since the values taken by $\boldsymbol{\mathcal{X}}$ are close to 0, the result of $\boldsymbol{\mathcal{X}}^2(0.35)\boldsymbol{\mathcal{X}}(0.85)$ is a very small number. So, its effect over $\Upsilon_{\mathrm{d}}\left(\boldsymbol{\mathcal{X}}\right)$ is expected to be negligible. As can be observed in the aforementioned figure, the values around $t\in \{0.35, 0.85\}$ have almost no impact, although there is a local maximum at $t=0.85$. Regarding $t=0.55$, as the absolute value is applied to $\boldsymbol{\mathcal{X}}(0.55)$, the degree of variability of $\boldsymbol{\mathcal{X}}(0.55)$ is limited to half. Therefore, the relevance of the point $t=0.55$ is expected to be low. These conclusions are supported by the Shapley value relevance functions depicted in the aforementioned figure.

The second column of Figure \ref{fig:shapley_scenarios_m_200_eta_005} corresponds to the scenarios where data are generated using the symmetric Fourier expansion. The same insights derived for the first column can be applied in these 4 scenarios, with the exception that, as symmetric functions are employed, in all cases two relevant areas are obtained, which are symmetric with respect to the midpoint $t=0.5$.

Finally, let us consider the last column of Figure \ref{fig:shapley_scenarios_m_200_eta_005}, which corresponds to scenarios whose data are generated using a Brownian motion with a trend. The first row corresponds to a linear unimodal transformation. All Shapley value relevance functions consider the most relevant point to be the one that maximizes $\beta_u(t)$. In the case of the scenario where a linear bimodal is used (second row), two areas are particularly relevant in this regard: the first corresponds with points close to $t_1=0.25$ and the second with points close to $t_2=0.75$, being $\{t_1, t_2\} =\argmax_{t \in I}\beta_b(t)$. As the variance of the Brownian trend increases with $t$, there is more variability for points close to $t=0.75$ than for points close to $t=0.25$, and therefore, it is expected to obtain a higher relevance in the neighborhood of $t_2$, a phenomenon that is indeed observed.

\begin{figure}
	\centering
	\includegraphics[width=.6\textwidth]{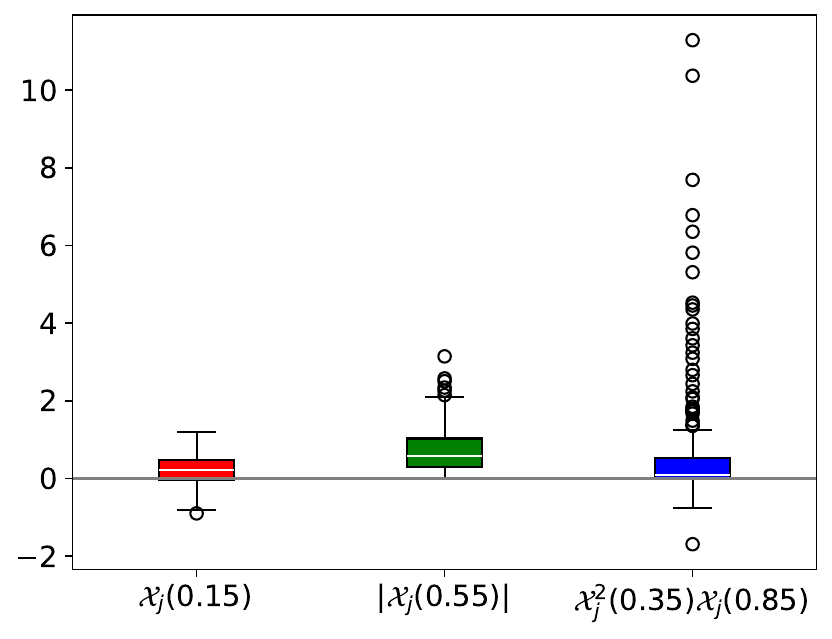}
	\caption{(Left) Boxplot of $\{ \mathcal{X}_j(0.15)\}_{j=1}^m$. (Middle) Boxplot of $\{ \rvert\mathcal{X}_j(0.55)\lvert\}_{j=1}^m$. (Right) Boxplot of $\{ \mathcal{X}_j^2(0.35) \mathcal{X}_j(0.85)\}_{j=1}^m$ where, $\{ \mathcal{X}_j \}_{j=1}^m$ corresponds to the Brownian trend data set shown in the rightmost panel of Figure \ref{fig:functional_covariates}.}
	\label{fig:boxplot_brownian}
\end{figure}

A similar argument explains why the highest relevance of the non-linear transformation (third row) is around $t=0.75$: as the variance increases with $t$, $\max_{t \in I} \left\{\lvert \beta_{\mathrm{u}}(t) \boldsymbol{\mathcal{X}}(t) \rvert,\right.$ $\left. \lvert \beta_{\mathrm{u}}(t) \boldsymbol{\mathcal{X}}(1 - t) \rvert \right\}$ is around $t=0.75$ instead of $t=0.25$. With regard to the discrete transformation (fourth row), the most relevant point is $t=0.35$, which is explained by (1) now the range of variability of $\boldsymbol{\mathcal{X}}^2(0.35)\boldsymbol{\mathcal{X}}(0.85)$ is larger than that of $\boldsymbol{\mathcal{X}}(0.15)$ and $\lvert \boldsymbol{\mathcal{X}}(0.55) \rvert$ (see Figure \ref{fig:boxplot_brownian}, corresponding to the Brownian trend data set) and (2) $\boldsymbol{\mathcal{X}}(0.35)$ is more difficult to predict than $\boldsymbol{\mathcal{X}}(0.85)$ using conditional expectations, because $t=0.85$ is closer to the end of the path $\{\boldsymbol{\mathcal{X}}(t):t\in[0,1]\}$ than $t=0.35$.

A final remark on the execution times is in order. The main computational burden of the algorithm is the large number of random permutations required in order to approximate the Shapley values. For instance, a single simulation of those employed to generate the graphic of first row and first column of Figure \ref{fig:shapley_scenarios_m_200_eta_005} (which uses 1000 random permutations) takes 25 seconds to compute the FLM Shapley value relevance function in an iMac (2024) with an M4 chip (10 cores; four performance cores and six efficiency cores running up to 4.4 GHz and 2.9 GHz respectively) and 16 GB of RAM memory. To evaluate the effect of the number of random permutations, we repeat this experiment with 2000 and 5000 random permutations, resulting in computation times of 48 and 109 seconds respectively.

\subsection{Real data}
\label{sec:exp_real_data}

\begin{figure}
	\centering
	
	\begin{subfigure}[t]{.32\textwidth}
		\centering
		\includegraphics[width=\textwidth]{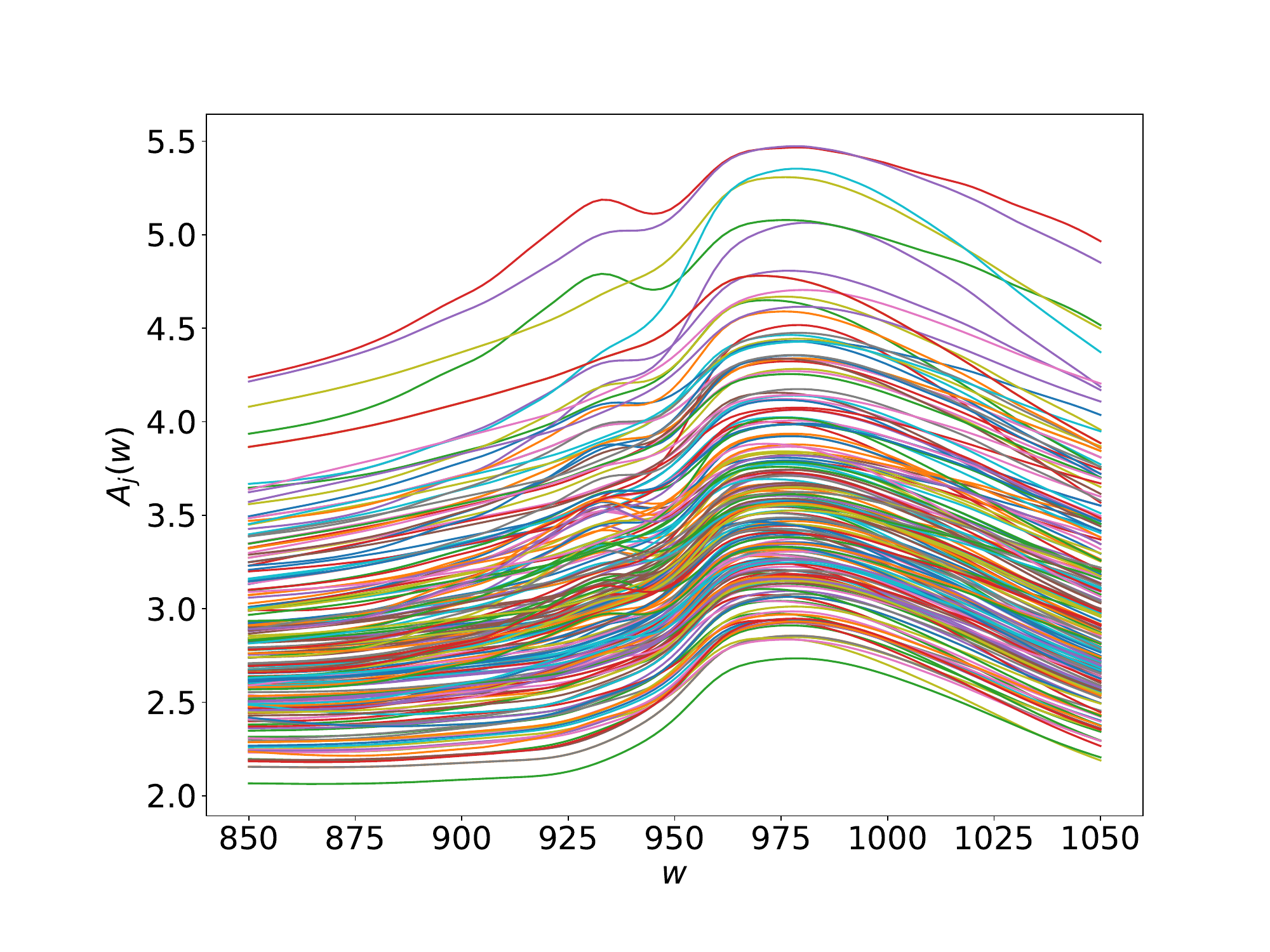}
		\caption{Spectometric curves of Tecator.}
	\end{subfigure}
	\hfill
	\begin{subfigure}[t]{.32\textwidth}
		\centering
		\includegraphics[width=\textwidth]{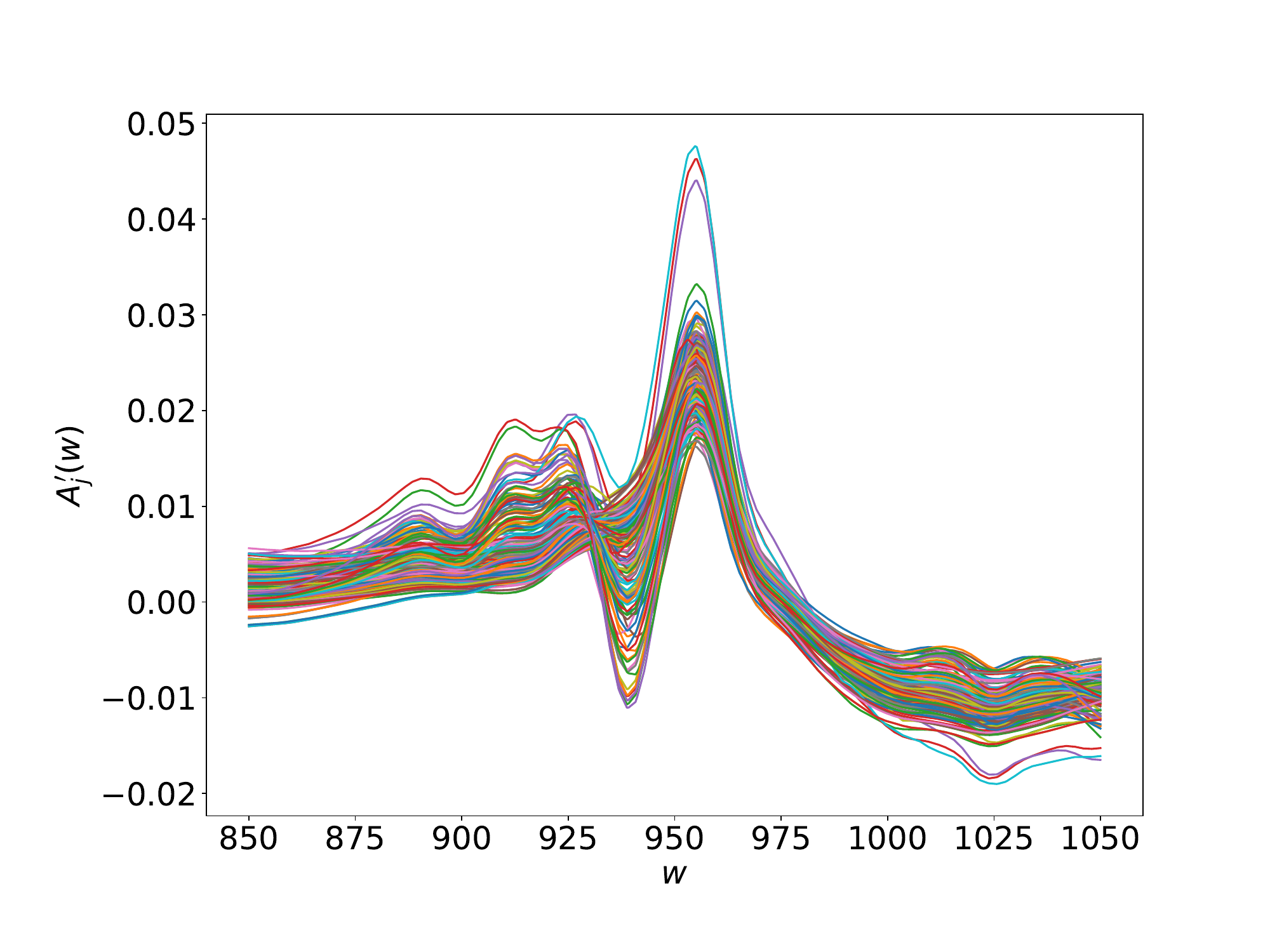}
		\caption{First derivative.}
	\end{subfigure}
	\hfill
	\begin{subfigure}[t]{.32\textwidth}
		\centering
		\includegraphics[width=\textwidth]{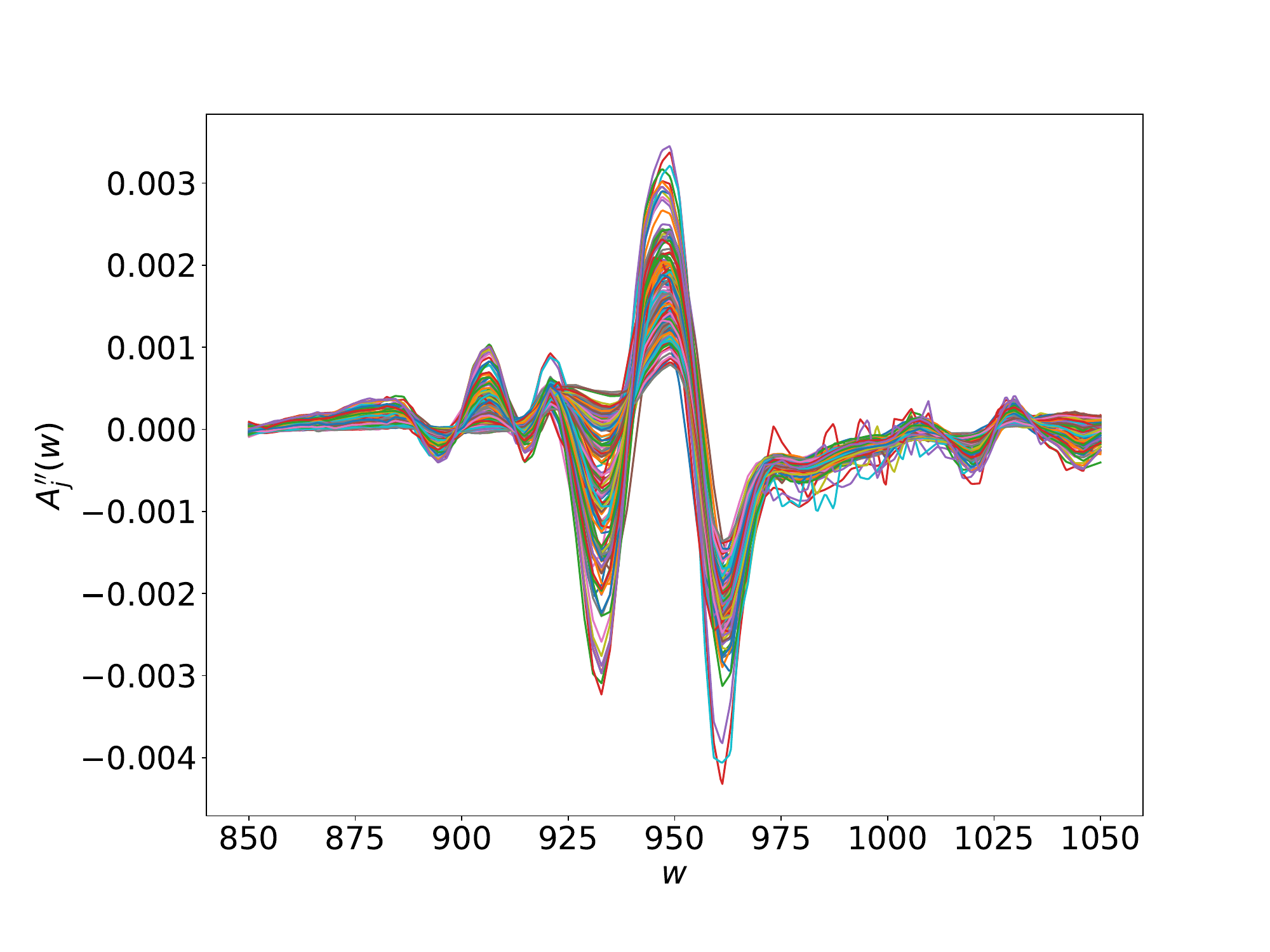}
		\caption{Second derivative.}
	\end{subfigure}
	
	\caption{Tecator data set.}
	\label{fig:tecator_raw}
\end{figure}

The Tecator data set has been widely used within the FDA literature (see, for instance, \citealp{Ferraty2006}). This data set consists of 215 spectrometric curves of meat samples and it measures the near infrared (NIR) absorbance $\mathcal{A}(w)$ as a function of wavelength $w$ (in nanometers). In addition, each curve is associated with three distinct quantities: the percentage of fat, water, and protein in the meat sample. The percentage of fat is used here as the target variable. Figure \ref{fig:tecator_raw} depicts the set of spectrometric curves (left), their first derivatives (middle) and their second derivatives (right).

Following \citet[Section 7.2.1]{Ferraty2006}, we divide the data set into {\em learning sample} (curves 1 to 160; $74.4\%$) and {\em testing sample} (curves 161 to 215; 55 curves, $25.6\%$). Given that we differentiate between {\em training} and {\em validation} sets, we divide the learning sample randomly into {\em training set} (128 curves; $59.6\%$) and {\em validation set} (32 curves; $14.8\%$). The training set is employed for fitting the same prediction algorithms used in Section \ref{sec:exp_simulated_data}: FLM, FKNN and FNN. The validation set is used to perform hyperparameter tuning, and the test data set serves to compute the distinct Shapley value relevance functions.

\begin{figure}
	\centering
	\begin{subfigure}{\textwidth}
        \centering
        \includegraphics[height=.29\textheight]{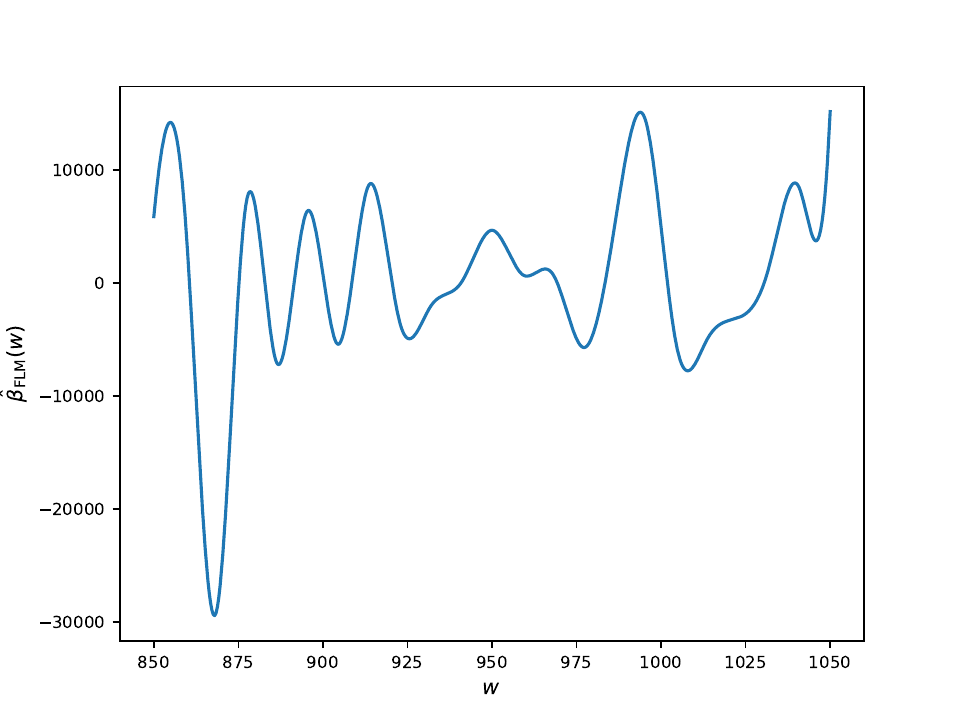}
    \end{subfigure}

    \begin{subfigure}{\textwidth}
        \centering
        \includegraphics[height=.29\textheight]{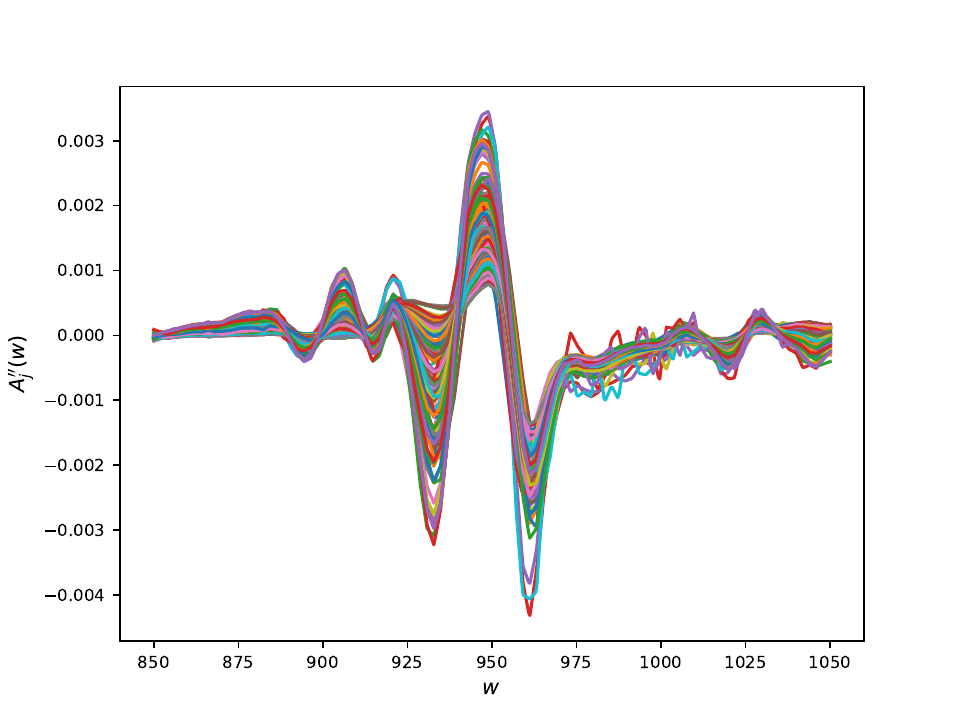}
    \end{subfigure}

	\begin{subfigure}{\textwidth}
		\centering
		\includegraphics[height=.29\textheight]{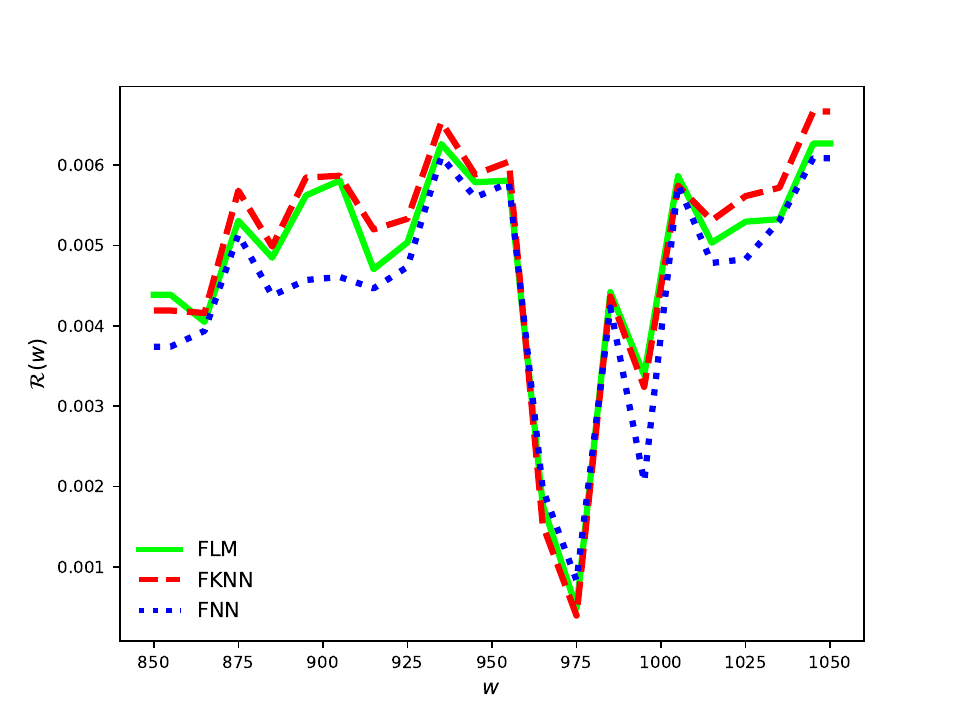}
	\end{subfigure}

	\caption{Shapley value analysis of Tecator data. (Top) Estimated beta function for the functional linear model using Tecator data set to predict the fat content. (Middle) Set of second derivatives. (Bottom) Shapley value relevance functions.}
	\label{fig:beta_flm_tecator_shapley}
\end{figure}

The prediction algorithms are used to model the target variable as a function of the second derivatives of the spectrometric curves, as suggested in \citet[Section 7.2.2]{Ferraty2006}. See also \citet{Boj2010}. We proceed in the same way as in Section \ref{sec:exp_simulated_data} to select the basis to represent the functional data. The $R^2(I)$ are 0.962 for FLM, 0.9808 for FKNN and 0.8196 for FNN. Regarding FLM, the top panel of Figure \ref{fig:beta_flm_tecator_shapley} shows the estimated function $\hat{\beta}(t)$. This function is very variable and difficult to interpret, since its shape does not seem to be related to that of the second derivatives (middle panel). Therefore, even FLM can benefit from the definition of a relevance function.

Next, we use our proposal to obtain interpretability for each of the prediction models, as in Section \ref{sec:exp_simulated_data}. The interval of interest, $I = [850, 1050]$, is divided into 20 parts, all of the same length, and 5000 permutations are performed. The bottom panel of Figure \ref{fig:beta_flm_tecator_shapley} depicts the Shapley value relevance functions, aligned with the set of second derivatives (middle panel). It should be recalled that the relevance is obtained at interval level. The three Shapley value relevance functions show their global maximum at $[1040, 1050]$. In this interval, the set of second derivatives manifests a certain degree of variability. On the other hand, all the Shapley value relevance functions consider the region within the interval $[930, 960)$ to be relevant. In this interval, the set of second derivatives exhibits rapid fluctuations between minimum and maximum, followed by another minimum. Finally, it is worth noting that all Shapley value relevance functions identify the interval $[970, 980)$ as the least relevant. This is likely due to the fact that the values of all second derivatives are close to each other at this interval.

\section{Conclusions}
\label{sec:conclusions}

This work represents a novel approach to addressing the global agnostic interpretability of prediction models within the field of functional data analysis. Our framework is based on game theory for games with a continuum of players, extending the Shapley value to the case of a set with infinite regressors. The central piece of this work is the Shapley value relevance function, that measures the relevance of all points of the interval where the functional data are observed. Alongside this manuscript, we present an open source \proglang{Python} package that implements this framework.

We have illustrated the performance of our work by means of a set of experiments. On the one hand, we use simulated data where the relevant points were known beforehand. The results have shown that these points have been successfully captured by the explainability method. On the other hand, we have explored the Tecator data set, where the second derivative has been used to model the data. According to our proposal, the most relevant intervals coincide with large variability areas of the second derivatives.

As future research, it would be interesting to extend this methodology in two directions: first, considering several functional regressors, and second, allowing a functional response and regressors that are either scalar or functional.

\appendix

\section{Interpretable machine learning}
\label{sec:iml}

This appendix summarizes the topic of interpretable prediction in the context of multiple regression. For a more detailed coverage see, for instance, \cite{Biecek2021} or \cite{Molnar2022}.

\cite{Breiman2001} identifies two distinct cultures within the field of data analysis: the modeling culture, based on statistical inference, and the prediction culture, focused on machine learning techniques. In addition, \cite{Breiman2001} points out a potential trade-off between predictive ability and interpretability of models. The predictive accuracy of machine learning algorithms, such as neural networks or random forests, often exceeds that of statistical models, such as linear or logistic regression. However, statistical models provide a simpler understanding of the relationship between the response variable and the input variables. In fact, machine learning algorithms are often referred to as ``black boxes'' because of their inability to provide understandable explanations of the reasons behind their predictions. Nevertheless, \cite{Breiman2001} calls for the development of procedures that would allow for better interpretation of algorithmic models without compromising their predictive ability.

In interpretable machine learning, a distinction is done between {\em global} and {\em local} interpretability. On the one hand, global interpretability tools measure the importance or relevance of each explanatory variable in the prediction process over their whole support. On the other hand, local interpretability tools provide meaningful explanations of why the prediction model returns a certain estimated response for a given individual, identified with a particular combination of the values of the predictor variables.

It is also relevant to classify interpretability methods as {\em model-specific} or {\em model-agnostic}. The first category includes interpretability methods that are designed to interpret a particular prediction algorithm (such as random forests or neural networks), exploiting its internal structure for interpretation. In contrast, model-agnostic interpretability methods can be applied to any prediction model. They only need to evaluate the prediction model on data from the training or test sets, or perturbations of either, and do not have access to the internal structure of the prediction model.

In this paper, we are interested in global model-agnostic interpretability methods. In the usual framework of a prediction problem with $p$ explanatory variables and one response to which a prediction model is fitted, global model-agnostic methods provide a relevance measure for each predictor. The most natural way to do this is {\em leave-one-covariate-out} (LOCO), and it has been used in multiple linear regression for decades. Two prediction models are fitted (one with all the predictors and the other with one predictor omitted) and their prediction errors are compared: the more different they are, the more important the omitted predictor is. The main drawback of LOCO is that it requires fitting $p$ additional models, each with $(p-1)$ predictors. 

In \cite{Breiman2001}, a permutation-based approach is used to define variable importance in random forests, which is easily extended to any prediction model when a test sample is available. The average prediction error for the test sample is compared with the same measure when the test sample is modified randomly permuting the values of a specific explanatory variable. The larger that difference, the more important is the permuted variable. Despite its popularity, using random permutations for interpreting black box prediction algorithms has received numerous criticisms (see, for instance, \citealp{Hooker2021,Delicado2023}), mainly when the predictors are moderate or highly correlated. 

\cite{Delicado2023} propose an alternative approach. First, they fit the model with all the explanatory variables in the training sample. Then, they measure the individual relevance of each explanatory variable by comparing the predictions of the model in the test set with those obtained when an explanatory variable in the test set is replaced by its \textit{ghost variable}, defined as the conditional expectation of that variable given the values of the other explanatory variables (using a linear model). \cite{Delicado2023} show that, in linear models, the proposed measure gives results similar to LOCO and outperforms random permutations.

A different line of work in global model-agnostic relevance measure is based on the Shapley value, a concept coming from game theory. We devote Section \ref{sec:game_theory} entirely to this approach, which is the one we follow in Section \ref{sec:shapley_fda} to develop our proposal.

As mentioned in the main corpus of the paper, we only found two papers considering interpretability in functional data regression models. \cite{James2009} specifically consider the functional linear regression model, whose global interpretation is favored when the estimate of $\beta(t)$ is exactly zero over regions with no apparent relationship between the functional predictor and the scalar response. The authors propose an estimation method which combines the basis expansion of $\beta(t)$ and a lasso-type strategy for variable selection. \cite{Carrizosa2024} address the issue of local feature importance using counterfactual analysis for functional data. So, their goal is to explain (or rank) the features in a neighborhood of a given individual, with functional observations $\mathcal{X}_0(t)$. The method is presented in the context of (multiclass) classification. The work aims at obtaining a counterfactual explanation $\mathcal{X}(t)$, which is an artificial functional data, by combining existing instances in the dataset, called {\em prototypes}. To achieve it, an optimization problem is formulated, taking into account that the cost of perturbing $\mathcal{X}_0(t)$ to obtain $\mathcal{X}(t)$ must be minimal.

\section{Additional results of the simulation study}
\label{sec:results_simulation_study}

In this section we present the results of the remaining scenarios, which are:
\begin{itemize}
	\item $m=200$ and $\eta=0.25$. See Figure \ref{fig:shapley_scenarios_m_200_eta_025} and Table \ref{table:r2_m_200_eta_025}.
	\item $m=500$ and $\eta=0.05$. See Figure \ref{fig:shapley_scenarios_m_500_eta_005} and Table \ref{table:r2_m_500_eta_005}.
	\item $m=500$ and $\eta=0.25$. See Figure \ref{fig:shapley_scenarios_m_500_eta_025} and Table \ref{table:r2_m_500_eta_025}.
\end{itemize}

\begin{figure}[hp]
	\centering
	
	\generategridfigures{03__fourier_expansion__linear_unimodal__200__025,23__symmetric_fourier_expansion__linear_unimodal__200__025,43__brownian_trend__linear_unimodal__200__025,07__fourier_expansion__linear_bimodal__200__025,27__symmetric_fourier_expansion__linear_bimodal__200__025,47__brownian_trend__linear_bimodal__200__025,11__fourier_expansion__non_linear_unimodal__200__025,31__symmetric_fourier_expansion__non_linear_unimodal__200__025,51__brownian_trend__non_linear_unimodal__200__025,19__fourier_expansion__linear_discrete__200__025,39__symmetric_fourier_expansion__linear_discrete__200__025,59__brownian_trend__linear_discrete__200__025}
	
	\includegraphics[width=.4\textwidth]{legend.pdf}
	\caption{Mean Shapley value relevance functions for those scenarios with $m=200$ and $\eta=0.25$.}
	\label{fig:shapley_scenarios_m_200_eta_025}
\end{figure}

\begin{table}[h]
	\centering
	
	\generategridtable
	{0.7317 (0.0282),0.7399 (0.0260),0.7384 (0.0227),0.3987 (0.0412),0.5235 (0.0415),0.6429 (0.0322),0.6936 (0.0570),0.7137 (0.0418),0.6884 (0.0655)}
	{0.7315 (0.0278),0.7358 (0.0303),0.7441 (0.0257),0.3995 (0.0398),0.5192 (0.0355),0.7061 (0.0280),0.6542 (0.0720),0.6945 (0.0683),0.6919 (0.0586)}
	{0.0271 (0.0285),0.0206 (0.0225),0.4460 (0.0798),0.0184 (0.0692),0.2036 (0.0551),0.6566 (0.0342),0.2147 (0.0952),0.5086 (0.1176),0.6271 (0.0864)}
	{0.5010 (0.0556),0.5248 (0.0488),0.4911 (0.0490),0.3161 (0.0513),0.4414 (0.0395),0.6274 (0.0530),0.4504 (0.1057),0.4973 (0.0821),0.5677 (0.0618)}

	\caption{Mean value (standard deviation) of $R^2(I)$ of FLM, FKNN and FNN for those scenarios with $m=200$ and $\eta=0.25$.}
	\label{table:r2_m_200_eta_025}
\end{table}

\begin{figure}[hp]
	\centering
	
	\generategridfigures{02__fourier_expansion__linear_unimodal__500__005,22__symmetric_fourier_expansion__linear_unimodal__500__005,42__brownian_trend__linear_unimodal__500__005,06__fourier_expansion__linear_bimodal__500__005,26__symmetric_fourier_expansion__linear_bimodal__500__005,46__brownian_trend__linear_bimodal__500__005,10__fourier_expansion__non_linear_unimodal__500__005,30__symmetric_fourier_expansion__non_linear_unimodal__500__005,50__brownian_trend__non_linear_unimodal__500__005,18__fourier_expansion__linear_discrete__500__005,38__symmetric_fourier_expansion__linear_discrete__500__005,58__brownian_trend__linear_discrete__500__005}
	
	\includegraphics[width=.4\textwidth]{legend.pdf}
	\caption{Mean Shapley value relevance functions for those scenarios with $m=500$ and $\eta=0.05$.}
	\label{fig:shapley_scenarios_m_500_eta_005}
\end{figure}

\begin{table}[h]
	\centering
	
	\generategridtable
	{0.9488 (0.0034),0.9491 (0.0034),0.9484 (0.0032),0.6044 (0.0274),0.7550 (0.0165),0.8805 (0.0078),0.9281 (0.0573),0.9433 (0.0136),0.9006 (0.0636)}
	{0.9486 (0.0031),0.9491 (0.0035),0.9492 (0.0034),0.6046 (0.0203),0.7556 (0.0134),0.9258 (0.0052),0.8963 (0.0917),0.9329 (0.0616),0.9078 (0.0463)}
	{0.0111 (0.0113),0.0108 (0.0116),0.5820 (0.0531),0.1108 (0.0437),0.4303 (0.0330),0.8828 (0.0093),0.5266 (0.0978),0.8049 (0.0548),0.8715 (0.0384)}
	{0.6872 (0.0252),0.6931 (0.0242),0.6259 (0.0322),0.4973 (0.0279),0.6640 (0.0183),0.8615 (0.0273),0.7136 (0.0841),0.7197 (0.0705),0.7506 (0.0477)}

	\caption{Mean value (standard deviation) of $R^2(I)$ of FLM, FKNN and FNN for those scenarios with $m=500$ and $\eta=0.05$.}
	\label{table:r2_m_500_eta_005}
\end{table}

\begin{figure}[hp]
	\centering
	
	\generategridfigures{04__fourier_expansion__linear_unimodal__500__025,24__symmetric_fourier_expansion__linear_unimodal__500__025,44__brownian_trend__linear_unimodal__500__025,08__fourier_expansion__linear_bimodal__500__025,28__symmetric_fourier_expansion__linear_bimodal__500__025,48__brownian_trend__linear_bimodal__500__025,12__fourier_expansion__non_linear_unimodal__500__025,32__symmetric_fourier_expansion__non_linear_unimodal__500__025,52__brownian_trend__non_linear_unimodal__500__025,20__fourier_expansion__linear_discrete__500__025,40__symmetric_fourier_expansion__linear_discrete__500__025,60__brownian_trend__linear_discrete__500__025}
	
	\includegraphics[width=.4\textwidth]{legend.pdf}
	\caption{Mean Shapley value relevance functions for those scenarios with $m=500$ and $\eta=0.25$.}
	\label{fig:shapley_scenarios_m_500_eta_025}
\end{figure}

\begin{table}[h!]
	\centering
	
	\generategridtable
	{0.7432 (0.0180),0.7476 (0.0134),0.7435 (0.0151),0.4581 (0.0221),0.5741 (0.0195),0.6728 (0.0191),0.7247 (0.0513),0.7300 (0.0427),0.7122 (0.0458)}
	{0.7415 (0.0185),0.7442 (0.0172),0.7430 (0.0157),0.4613 (0.0252),0.5719 (0.0225),0.7164 (0.0168),0.6979 (0.0637),0.7253 (0.0564),0.7165 (0.0357)}
	{0.0102 (0.0106),0.0107 (0.0114),0.4560 (0.0457),0.0641 (0.0463),0.2988 (0.0350),0.6799 (0.0176),0.3975 (0.0758),0.6285 (0.0463),0.6788 (0.0369)}
	{0.5387 (0.0284),0.5385 (0.0277),0.4838 (0.0312),0.3739 (0.0303),0.5049 (0.0237),0.6500 (0.0308),0.5400 (0.0697),0.5731 (0.0503),0.5929 (0.0391)}

	\caption{Mean value (standard deviation) of $R^2(I)$ of FLM, FKNN and FNN for those scenarios with $m=500$ and $\eta=0.25$.}
	\label{table:r2_m_500_eta_025}
\end{table}

\newpage

\section*{Acknowledgments}
{\small This research was supported by projects PID2020-116294GB-I00 and PID2023-148158OB-I00 (founded by MICIU/AEI/10.13039/501100011033 and European Union NextGeneratio-EU/PRTR), by AGAUR under grants 2020 FI SDUR 306 and 2021 SGR 00613, and by UPC under AGRUPS-2024.}

\bibliography{references}

\begin{thebibliography}{40}
\providecommand{\natexlab}[1]{#1}
\providecommand{\url}[1]{\texttt{#1}}
\expandafter\ifx\csname urlstyle\endcsname\relax
  \providecommand{\doi}[1]{doi: #1}\else
  \providecommand{\doi}{doi: \begingroup \urlstyle{rm}\Url}\fi

\bibitem[Aumann and Shapley(1974)]{Aumann1974}
R.~J. Aumann and L.~S. Shapley.
\newblock \emph{{Values of Non-Atomic Games}}.
\newblock Princeton University Press, 1974.
\newblock ISBN 978-0-691-64546-9.

\bibitem[Aumann(1964)]{Aumann1964}
Robert~J. Aumann.
\newblock {Markets with a Continuum of Traders}.
\newblock \emph{Econometrica}, 32:\penalty0 39--50, 1964.
\newblock \doi{10.2307/1913732}.

\bibitem[{Barredo Arrieta} et~al.(2020){Barredo Arrieta}, Díaz-Rodríguez, Ser, Bennetot, Tabik, Barbado, Garcia, et~al.]{Barredo2020}
Alejandro {Barredo Arrieta}, Natalia Díaz-Rodríguez, Javier~Del Ser, Adrien Bennetot, Siham Tabik, Alberto Barbado, Salvador Garcia, et~al.
\newblock {Explainable Artificial Intelligence (XAI): Concepts, taxonomies, opportunities and challenges toward responsible AI}.
\newblock \emph{Information Fusion}, 58:\penalty0 82--115, 2020.
\newblock \doi{10.1016/j.inffus.2019.12.012}.

\bibitem[Biecek and Burzykowski(2021)]{Biecek2021}
P~Biecek and T~Burzykowski.
\newblock \emph{{Explanatory model analysis: Explore, explain and examine predictive models}}.
\newblock Chapman and Hall/CRC, 2021.
\newblock ISBN 978-0-367-13559-1.

\bibitem[Boj et~al.(2010)Boj, Delicado, and Fortiana]{Boj2010}
Eva Boj, Pedro Delicado, and Josep Fortiana.
\newblock {Distance-based local linear regression for functional predictors}.
\newblock \emph{Computational Statistics \& Data Analysis}, 54\penalty0 (2):\penalty0 429--437, 2010.
\newblock \doi{10.1016/j.csda.2009.09.010}.

\bibitem[Borggaard and Thodberg(1992)]{Borggaard1992}
Claus Borggaard and Hans~Henrik Thodberg.
\newblock {Optimal minimal neural interpretation of spectra}.
\newblock \emph{Analytical Chemistry}, 64\penalty0 (5):\penalty0 545--551, 1992.
\newblock \doi{10.1021/ac00029a018}.

\bibitem[Breiman(2001)]{Breiman2001}
Leo Breiman.
\newblock {Statistical Modeling: The Two Cultures}.
\newblock \emph{Statistical Science}, 16\penalty0 (3):\penalty0 199--231, 2001.
\newblock \doi{10.1214/ss/1009213726}.

\bibitem[Carrizosa et~al.(2024)Carrizosa, Ramírez-Ayerbe, and Romero~Morales]{Carrizosa2024}
Emilio Carrizosa, Jasone Ramírez-Ayerbe, and Dolores Romero~Morales.
\newblock {A new model for counterfactual analysis for functional data}.
\newblock \emph{Advances in Data Analysis and Classification}, 18\penalty0 (4):\penalty0 981--1000, 2024.
\newblock \doi{10.1007/s11634-023-00563-5}.

\bibitem[Cohen et~al.(2007)Cohen, Dror, and Ruppin]{Cohen2007}
Shay~B. Cohen, Gideon Dror, and Eytan Ruppin.
\newblock {Feature Selection via Coalitional Game Theory}.
\newblock \emph{Neural Computation}, 19\penalty0 (7):\penalty0 1939--1961, 2007.
\newblock \doi{10.1162/neco.2007.19.7.1939}.

\bibitem[Crainiceanu et~al.(2024)Crainiceanu, Goldsmith, Leroux, and Cui]{Ciprian2024}
Ciprian~M. Crainiceanu, Jeff Goldsmith, Andrew Leroux, and Erjia Cui.
\newblock \emph{{Functional Data Analysis with R}}.
\newblock Chapman and Hall/CRC, 2024.
\newblock ISBN 978-1-032-24472-3.

\bibitem[Delicado and Peña(2023)]{Delicado2023}
Pedro Delicado and Daniel Peña.
\newblock {Understanding complex predictive models with ghost variables}.
\newblock \emph{TEST}, 32\penalty0 (1):\penalty0 107--145, 2023.
\newblock \doi{10.1007/s11749-022-00826-x}.

\bibitem[Farebrother(1988)]{Farebrother1988}
R.~W. Farebrother.
\newblock \emph{{Linear least squares computations}}.
\newblock Routledge, 1988.
\newblock ISBN 978-0-824-77661-9.
\newblock \doi{10.1201/9780203748923}.

\bibitem[Feldman(2005)]{Feldman2005}
Barry~E. Feldman.
\newblock {Relative Importance and Value}.
\newblock Available at SSRN Electronic Journal, doi: 10.2139/ssrn.2255827, 2005.

\bibitem[Ferraty and Vieu(2006)]{Ferraty2006}
F.~Ferraty and P.~Vieu.
\newblock \emph{{Nonparametric Functional Data Analysis: Theory and Practice}}.
\newblock Springer, 2006.
\newblock ISBN 978-0-387-30369-7.

\bibitem[Gertheiss et~al.(2024)Gertheiss, Rügamer, Liew, and Greven]{Gertheiss2024}
Jan Gertheiss, David Rügamer, Bernard X.~W. Liew, and Sonja Greven.
\newblock {Functional Data Analysis: An Introduction and Recent Developments}.
\newblock \emph{Biometrical Journal}, 66\penalty0 (7):\penalty0 e202300363, 2024.
\newblock \doi{10.1002/bimj.202300363}.

\bibitem[Goldberg et~al.(2014)Goldberg, Ritov, and Mandelbaum]{Goldberg2014}
Y.~Goldberg, Y.~Ritov, and A.~Mandelbaum.
\newblock {Predicting the continuation of a function with applications to call center data}.
\newblock \emph{Journal of Statistical Planning and Inference}, 147:\penalty0 53--65, 2014.
\newblock \doi{10.1016/j.jspi.2013.11.006}.

\bibitem[Grömping(2009)]{Gromping2009}
Ulrike Grömping.
\newblock {Variable Importance Assessment in Regression: Linear Regression versus Random Forest}.
\newblock \emph{The American Statistician}, 63\penalty0 (4):\penalty0 308--319, 2009.
\newblock \doi{10.1198/tast.2009.08199}.

\bibitem[Harris et~al.(2020)Harris, Millman, van~der Walt, Gommers, Virtanen, Cournapeau, et~al.]{Harris2020}
Charles~R. Harris, K.~Jarrod Millman, Stéfan~J. van~der Walt, Ralf Gommers, Pauli Virtanen, David Cournapeau, et~al.
\newblock {Array programming with NumPy}.
\newblock \emph{Nature}, 585\penalty0 (7825):\penalty0 357--362, 2020.
\newblock \doi{10.1038/s41586-020-2649-2}.

\bibitem[Hart and Neyman(1988)]{Hart1988}
Sergiu Hart and Abraham Neyman.
\newblock {Values of non-atomic vector measure games: Are they linear combinations of the measures?}
\newblock \emph{Journal of Mathematical Economics}, 17\penalty0 (1):\penalty0 31--40, 1988.
\newblock \doi{10.1016/0304-4068(88)90025-0}.

\bibitem[Heinrichs et~al.(2023)Heinrichs, Heim, and Weber]{Heinrichs2023}
Florian Heinrichs, Mavin Heim, and Corinna Weber.
\newblock {Functional neural networks: Shift invariant models for functional data with applications to EEG classification}.
\newblock In \emph{{Proceedings of the 40th International Conference on Machine Learning}}, pages 12866--12881. PMLR, 2023.

\bibitem[Hooker et~al.(2021)Hooker, Mentch, and Zhou]{Hooker2021}
G.~Hooker, L.~Mentch, and S.~Zhou.
\newblock {Unrestricted permutation forces extrapolation: variable importance requires at least one more model, or there is no free variable importance}.
\newblock \emph{Statistics and Computing}, 31\penalty0 (6):\penalty0 82, 2021.
\newblock \doi{10.1007/s11222-021-10057-z}.

\bibitem[Horváth and Kokoszka(2012)]{Horvath2012}
L.~Horváth and P.~Kokoszka.
\newblock \emph{{Inference for Functional Data with Applications}}.
\newblock Springer, 2012.
\newblock ISBN 978-1-4614-3654-6.

\bibitem[James et~al.(2009)James, Wang, and Zhu]{James2009}
Gareth~M. James, Jing Wang, and Ji~Zhu.
\newblock {Functional linear regression that's interpretable}.
\newblock \emph{The Annals of Statistics}, 37\penalty0 (5A):\penalty0 2083--2108, 2009.
\newblock \doi{10.1214/08-AOS641}.

\bibitem[Kannai(1966)]{Kannai1966}
Yakar Kannai.
\newblock {Values of games with a continuum of players}.
\newblock \emph{Israel Journal of Mathematics}, 4\penalty0 (1):\penalty0 54--58, 1966.
\newblock \doi{10.1007/BF02760070}.

\bibitem[Kneip and Liebl(2020)]{Kneip2020}
Alois Kneip and Dominik Liebl.
\newblock {On the optimal reconstruction of partially observed functional data}.
\newblock \emph{The Annals of Statistics}, 48\penalty0 (3):\penalty0 1692--1717, 2020.
\newblock \doi{10.1214/19-AOS1864}.

\bibitem[Kokoszka and Reimherr(2017)]{Kokoszka2017}
P.~Kokoszka and M.~Reimherr.
\newblock \emph{{Introduction to Functional Data Analysis}}.
\newblock Chapman \& Hall/CRC, 2017.
\newblock ISBN 978-1-498-74634-2.

\bibitem[Kraus(2015)]{Kraus2015}
David Kraus.
\newblock {Components and completion of partially observed functional data}.
\newblock \emph{Journal of the Royal Statistical Society: Series B (Statistical Methodology)}, 77\penalty0 (4):\penalty0 777--801, 2015.
\newblock \doi{10.1111/rssb.12087}.

\bibitem[Lipovetsky and Conklin(2001)]{Lipovetsky2001}
Stan Lipovetsky and Michael Conklin.
\newblock {Analysis of regression in game theory approach}.
\newblock \emph{Applied Stochastic Models in Business and Industry}, 17\penalty0 (4):\penalty0 319--330, 2001.
\newblock \doi{10.1002/asmb.446}.

\bibitem[Masís(2021)]{Masis2021}
Serg Masís.
\newblock \emph{{Interpretable Machine Learning with Python: Learn to build interpretable high-performance models with hands-on real-world examples}}.
\newblock Packt Publishing, 2021.
\newblock ISBN 978-1800203907.

\bibitem[Molnar(2022)]{Molnar2022}
Christoph Molnar.
\newblock \emph{{Interpretable Machine Learning: A Guide for Making Black Box Models Explainable}}.
\newblock Independently published, 2022.
\newblock ISBN 979-8-411-46333-0.

\bibitem[Neyman(1994)]{Neyman1994}
A.~Neyman.
\newblock \emph{{Value of Games with a Continuum of Players}}, chapter~VI, pages 67--79.
\newblock Springer, 1994.
\newblock ISBN 978-0-7923-3011-0.
\newblock \doi{10.1007/978-94-017-1656-7_7}.

\bibitem[{Python Core Team}(2021)]{PythonCoreTeam}
{Python Core Team}.
\newblock \emph{{The Python Language Reference}}.
\newblock Python Software Foundation, 2021.
\newblock URL \url{https://docs.python.org/3.10/reference/index.html}.

\bibitem[Ramos-Carreño et~al.(2024)Ramos-Carreño, Torrecilla, Carbajo-Berrocal, Marcos, and Suárez]{RamosCarreno2024}
Carlos Ramos-Carreño, José~Luis Torrecilla, Miguel Carbajo-Berrocal, Pablo Marcos, and Alberto Suárez.
\newblock {scikit-fda: A Python Package for Functional Data Analysis}.
\newblock \emph{Journal of Statistical Software}, 109\penalty0 (2):\penalty0 1--37, 2024.
\newblock \doi{10.18637/jss.v109.i02}.

\bibitem[Ramsay and Silverman(2005)]{Ramsay2005}
J.~Ramsay and B.W. Silverman.
\newblock \emph{{Functional Data Analysis}}.
\newblock Springer, 2005.
\newblock ISBN 978-0-387-40080-8.

\bibitem[Rao and Reimherr(2023)]{Aniruddha2023}
Aniruddha~Rajendra Rao and Matthew Reimherr.
\newblock {Nonlinear Functional Modeling Using Neural Networks}.
\newblock \emph{Journal of Computational and Graphical Statistics}, 32\penalty0 (4):\penalty0 1248--1257, 2023.
\newblock \doi{10.1080/10618600.2023.2165498}.

\bibitem[Reiss et~al.(2017)Reiss, Goldsmith, Shang, and Ogden]{Reiss:2017}
Philip~T. Reiss, Jeff Goldsmith, Han~Lin Shang, and R.~Todd Ogden.
\newblock {Methods for Scalar-on-Function Regression}.
\newblock \emph{International Statistical Review}, 85\penalty0 (2):\penalty0 228--249, 2017.
\newblock \doi{10.1111/insr.12163}.

\bibitem[Shapley(1953)]{Shapley1953}
Lloyd~S Shapley.
\newblock {A Value for n-Person Games}.
\newblock \emph{Contributions to the Theory of Games}, 2\penalty0 (28):\penalty0 307--317, 1953.
\newblock \doi{10.1515/9781400881970-018}.

\bibitem[Shapley(1961)]{Shapley1961}
Lloyd~S. Shapley.
\newblock {Values of Games With Infinitely Many Players}, 1961.
\newblock {\em Research memorandum.} RAND Corporation, doi: 10.7249/RM2912.

\bibitem[Winter(2002)]{Winter2002}
Eyal Winter.
\newblock {The Shapley Value}.
\newblock In Robert Aumann and S.~Hart, editors, \emph{{Handbook of Game Theory with Economic Applications}}, volume~3, chapter~53, pages 2025--2054. Elsevier, 2002.
\newblock ISBN 978-0-444-89428-1.
\newblock \doi{10.1016/S1574-0005(02)03016-3}.

\bibitem[Yao et~al.(2021)Yao, Mueller, and Wang]{Yao2021}
Junwen Yao, Jonas Mueller, and Jane-Ling Wang.
\newblock {Deep learning for functional data analysis with adaptive basis layers}.
\newblock In \emph{{International Conference on Machine Learning}}, pages 11898--11908. PMLR, 2021.

\end{thebibliography}

\end{document}